%% file: main.tex
\documentclass{article}

\usepackage{arxiv}

\usepackage[utf8]{inputenc} 
\usepackage[T1]{fontenc}    
\usepackage{hyperref}       
\usepackage{url}            
\usepackage{booktabs}       
\usepackage{amsfonts}       
\usepackage{nicefrac}       
\usepackage{microtype}      
\usepackage{graphicx}
\usepackage{doi}

\usepackage{mathptmx}
\usepackage{soul}\setuldepth{article}
\usepackage{csquotes} 
\usepackage{graphicx}
\usepackage{subfig} 
\usepackage{multirow}
\usepackage[table,xcdraw]{xcolor}

\def\hb{\hbox to 11.5 cm{}}

\begin{document}



\title{Mutual Understanding between People and Systems via Neurosymbolic AI and Knowledge Graphs} 

\author{\href{https://orcid.org/0000-0001-9962-7193}{\includegraphics[scale=0.06]{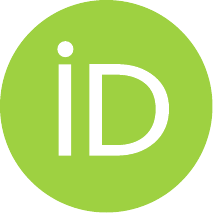}\hspace{1mm}Irene Celino}\\
Cefriel, viale Sarca 226, \\20126 Milano, Italy \\
\texttt{irene.celino@cefriel.com} 
\And 
\href{https://orcid.org/0000-0002-8235-7331}{\includegraphics[scale=0.06]{orcid.pdf}\hspace{1mm}Mario Scrocca}\\
Cefriel, viale Sarca 226, \\20126 Milano, Italy \\
\texttt{mario.scrocca@cefriel.com}   
\And 
\href{https://orcid.org/0000-0003-3594-731X}{\includegraphics[scale=0.06]{orcid.pdf}\hspace{1mm}Agnese Chiatti}\\
Politecnico di Milano, Piazza Leonardo da Vinci 32, \\20133 Milano, Italy \\
\texttt{agnese.chiatti@polimi.it}  
}

\date{}

\renewcommand{\headeright}{}
\renewcommand{\undertitle}{}
\renewcommand{\shorttitle}{I. Celino et al. -- Mutual Understanding between People and Systems via Neurosymbolic AI and Knowledge Graphs}


\maketitle

\begin{abstract}
This chapter investigates the concept of mutual understanding between humans and systems, positing that Neuro-symbolic Artificial Intelligence (NeSy AI) methods can significantly enhance this mutual understanding by leveraging explicit symbolic knowledge representations with data-driven learning models. We start by introducing three critical dimensions to characterize mutual understanding: sharing knowledge, exchanging knowledge, and governing knowledge. Sharing knowledge involves aligning the conceptual models of different agents to enable a shared understanding of the domain of interest. Exchanging knowledge relates to ensuring the effective and accurate communication between agents. Governing knowledge concerns establishing rules and processes to regulate the interaction between agents. Then, we present several different use case scenarios that demonstrate the application of NeSy AI and Knowledge Graphs to aid meaningful exchanges between human, artificial, and robotic agents. These scenarios highlight both the potential and the challenges of combining top-down symbolic reasoning with bottom-up neural learning, guiding the discussion of the coverage provided by current solutions along the dimensions of sharing, exchanging, and governing knowledge. Concurrently, this analysis facilitates the identification of gaps and less developed aspects in mutual understanding to address in future research.  \\

\end{abstract}

\section{Introduction}
\input{sezioni/intro}

\section{Dimensions of Mutual Understanding}\label{sec:framework}
\input{sezioni/dimensions}

\section{Case 1: Human Data Collection and Processing}\label{sec:coney-gwap}
\input{sezioni/coney-gwap}

\section{Case 2: Knowledge Extraction and Acquisition}
\input{sezioni/khub-perks}

\section{Case 3: Interoperability between Information Systems}\label{sec:sprint-tangent}
\input{sezioni/sprint-tangent}

\section{Case 4: Reproducibility and Trustability between Stakeholders}\label{sec:daydreams-ride2rail}
\input{sezioni/daydreams-ride2rail}

\section{Case 5: Building Shared Spatio-Temporal Understanding}\label{sec:spatiotemporal}
\input{sezioni/spatiotemporal-robots}

\section{Case 6: Hybrid Reasoning for Robot Deliberation}\label{sec:hybrid-reasoning}
\input{sezioni/hybrid-reasoning}

\section{Synthesis and Open Challenges for Mutual Understanding}\label{sec:challenges}

In this chapter, we summarised a series of efforts in knowledge graph and neurosymbolic AI and we analysed them with the lens of mutual understanding dimensions. The surveyed works touch very different aspects: the involved agents (human, digital and robotic), the types of knowledge (tacit vs. explicit, declarative vs. procedural vs. causal), the varying sources of such knowledge (people, data, documents, media), the type of AI techniques adopted (symbolic, neural, neurosymbolic). While we don't claim our analysis to be exhaustive, we covered a wide range of real-world cases, with their respective requirements and challenges, and our aspiration is that those concrete and pragmatic examples can be used as incentives for research directions and applied system designs to provide impact to a large set of stakeholders~\cite{chaves2024dagstuhlkg}.

\begin{table}[htb]
\caption{Summary of symbolic and neural approaches to solve the mutual understanding challenges (in the cells, darker color and H means higher coverage, lighter color and L means lower coverage).\vspace{.2cm}}
\label{tab:summary}
\centering
{\renewcommand{\arraystretch}{1.5}%
\begin{tabular}{|l|cc|cc|cc|}
\hline
\multicolumn{1}{|c|}{}                                    & \multicolumn{2}{c|}{Sharing}                                               & \multicolumn{2}{c|}{Exchanging}                                            & \multicolumn{2}{c|}{Governing}                                             \\ \cline{2-7} 
\multicolumn{1}{|c|}{\multirow{-2}{*}{Use Case Scenario}} & \multicolumn{1}{c|}{Symb}                      & Neur                      & \multicolumn{1}{c|}{Symb}                      & Neur                      & \multicolumn{1}{c|}{Symb}                      & Neur                      \\ \hline
Human Data Collection and Analysis                        & \multicolumn{1}{c|}{\cellcolor[HTML]{6D9EEB}H} & \cellcolor[HTML]{D9EAD3}L & \multicolumn{1}{c|}{\cellcolor[HTML]{C9DAF8}L} & \cellcolor[HTML]{D9EAD3}L & \multicolumn{1}{c|}{\cellcolor[HTML]{C9DAF8}L} &                           \\ \hline
Knowledge Extraction and Acquisition                      & \multicolumn{1}{c|}{\cellcolor[HTML]{6D9EEB}H} & \cellcolor[HTML]{93C47D}H & \multicolumn{1}{c|}{\cellcolor[HTML]{C9DAF8}L} & \cellcolor[HTML]{D9EAD3}L & \multicolumn{1}{c|}{\cellcolor[HTML]{C9DAF8}L} &                           \\ \hline
Interoperability between Information Systems                & \multicolumn{1}{c|}{\cellcolor[HTML]{C9DAF8}L} &                           & \multicolumn{1}{c|}{\cellcolor[HTML]{6D9EEB}H} & \cellcolor[HTML]{93C47D}H & \multicolumn{1}{c|}{\cellcolor[HTML]{C9DAF8}L} &                           \\ \hline
Reproducibility and Trustability                          & \multicolumn{1}{c|}{\cellcolor[HTML]{6D9EEB}H} & \cellcolor[HTML]{D9EAD3}L & \multicolumn{1}{c|}{}                          &                           & \multicolumn{1}{c|}{\cellcolor[HTML]{6D9EEB}H} & \cellcolor[HTML]{D9EAD3}L \\ \hline
Building Shared Spatio-Temporal Understanding             & \multicolumn{1}{c|}{\cellcolor[HTML]{6D9EEB}H} & \cellcolor[HTML]{93C47D}H & \multicolumn{1}{c|}{}                          & \cellcolor[HTML]{D9EAD3}L & \multicolumn{1}{c|}{}                          & \cellcolor[HTML]{D9EAD3}L \\ \hline
Hybrid Reasoning for Robot Deliberation                   & \multicolumn{1}{c|}{\cellcolor[HTML]{6D9EEB}H} & \cellcolor[HTML]{93C47D}H & \multicolumn{1}{c|}{\cellcolor[HTML]{6D9EEB}H} & \cellcolor[HTML]{93C47D}H & \multicolumn{1}{c|}{\cellcolor[HTML]{C9DAF8}L} &                           \\ \hline
\end{tabular}}
\end{table}

Table~\ref{tab:summary} graphically summarises the level of achievement of the three main dimensions of mutual understanding -- sharing, exchange and governance of knowledge -- across the presented scenarios.

Regarding the \emph{sharing knowledge dimension}, it is clear that existing solutions, especially in the symbolic AI area already well cover this challenge: conceptual models and structured data, especially when enriched with clear semantics, are the basis for a true mutual understanding between different agents. In this regards, neural approaches are already adopted for knowledge extraction, scene understanding and sense-making. There is still a clear potential for future evolution of knowledge sharing with mixed neurosymbolic AI solutions, which mix and match bottom-up with top-down semantic models of the domain and context to provide richer and more useful representations of the world.

We highlighted also several contributions of both symbolic and neural approaches to address the \emph{exchanging knowledge} challenge: while the interaction between different agents will always include some custom configuration and agreement to enable exchange, several directions are already visible. Data interoperability is a central ingredient to enable meaningful exchanges between systems and neurosymbolic approaches can definitely help in reducing the human costs by automating some steps. In particular, the advent of large-scale generative AI systems can support a new generation of conversational agents that enrich and support the communication between agents and, as such, the knowledge exchange.

The challenge that may require further attention and research is related to the \emph{governing knowledge dimension}: it has become mandatory to find solutions to the delicate task of managing processes, roles and rules to regulate the interactions between agents, especially in the era of the AI Act~\cite{aiact}. A number of symbolic approaches to address governance exist, especially by providing explicit representations of artifacts, business agreements and constraints; a few approaches also address the elicitation of implicit norms and patterns from data and background knowledge. We believe that the application of neurosymbolic AI to the governance challenge will result in further solutions to iteratively and collaboratively involve different stakeholders and agents in finding a mutual understanding on processes and rules to regulate their relationships and interactions.

\subsubsection*{Acknowledgments}
\small The authors would like to thank Antonia Azzini, Ilaria Baroni, Alessio Carenini, Valentina Carriero, Marco Comerio, Marco Grassi, Gloria Re Calegari, Enrico Motta, Gianluca Bardaro, Alessio Antonini, Enrico Daga for their precious collaboration in the research efforts summarised in this chapter. This work is partially supported by the PERKS project (Grant agreement ID: 101120323), co-funded by the European Commission under the Horizon Europe Framework Programme.

\normalsize

\bibliographystyle{plain}
\bibliography{references.bib}

\end{document}

%% file: sezioni/intro.tex
In his popular book ``Thinking, fast and slow"~\cite{kahneman2011thinking}, Daniel Kahneman proposed his  cognitive theory of human decision-making with a differentiation between two modes of thought -- a fast, instinctive and emotional system and a slower, more deliberative, and more logical one. The former is more unconscious and less explainable, showing some characteristics that resemble those of machine learning systems; the latter is more explicit and reasonable, displaying traits that are common in knowledge representation and reasoning systems.

When designing artificial intelligence systems, therefore, we can think of Neuro-symbolic AI~\cite{sarker2021neuro} as the combination of ``System 1" neural, connectionist approaches with ``System 2" symbolic intelligent systems, to build hybrid solutions that take the best out of the two worlds~\cite{booch2021thinking}.  Broadly speaking, Neuro-symbolic systems combine methods based on Artificial Neural Networks (i.e., the \textit{neural} component) with methods based on the explicit manipulation of \textit{symbols}. Although the distinction between symbols and non-symbols is a subject of active debate, we follow~\cite{van2021modular} in using the term symbol to refer to semantic descriptions of objects, classes and relations in the world (e.g., labels, relations, other explanations of historical data records). This hybrid learning process can also be seen as one that combines patterns extracted intuitively via bottom-up observation with top-down cognitive expectations about the environment \cite{cummings2021rethinking}.

With reference to human decision-making and inter-personal relationships, we are used to think that clear and meaningful conversations between people are needed to produce and achieve \emph{mutual understanding} and successful outcomes. If we refer to classical definitions like Wittgenstein's, we also know that such achievement is oftentimes hard to reach: ``We speak of understanding a sentence in the sense in which it can be replaced by another which says the same; but also in the sense in which it cannot be replaced by any other"~\cite{wittgenstein2009philosophical}, which challenges the very notion of understanding.

How can we promote achieve \emph{mutual understanding between people} when their interaction is mediated by a digital artificial system? Can we also foster \emph{mutual understanding between machines}, supporting interoperability and meaningful exchange? What are successful cases of employment of symbolic approaches (e.g., knowledge graphs and semantic web solutions), neural approaches (e.g., machine learning, generative AI, robotics) and a mix of them to achieve mutual understanding?

In this chapter, we aim at providing some answers to those questions, by proposing a set of dimensions to characterise mutual understanding (Section~\ref{sec:framework}), by illustrating and analysing different use case scenarios that adopt Neuro-symbolic AI and Knowledge Graphs (Sections~\ref{sec:coney-gwap}-\ref{sec:hybrid-reasoning}), and by providing a synthesis and challenges for future work (Section~\ref{sec:challenges}).

%% file: sezioni/dimensions.tex
\textbf{Mutual understanding} (between people) happens \enquote{when different minds mutually infer they agree on an understanding of an object, person, place, event, or idea} and this is usually supported by \textbf{conceptual alignment}, which is the \enquote{condition in which individuals’ mental representations have become aligned, or sufficiently compatible, despite those individuals’ idiosyncratic experiences and knowledge structures}~\cite{stolk2016conceptual}.

A classic study~\cite{tan1994establishing} on communication behaviours of human agents during analyst-client interviews highlighted that there are three key processes that affect mutual understanding between people: \emph{shifting perspective} (the process of tuning into the interlocutor's frame 
of reference \cite{tan1989investigation} or the individual's capacity to relate in a  manner most appropriate  to  the  particular  communication  situation \cite{montgomery1981form}), \emph{managing transactions} (the ability to handle the procedural aspects of structuring, controlling, and  maintaining  a conversation \cite{wiemann1977explication}), and \emph{establishing rapport} (an aspect of an interpersonal relationships that may be defined as creating a state of harmony, accordance,  and  congruity  developed  in  a  relationship  \cite{guinan1986development}).
Therefore, in the context of our investigation, we contextualise those three key processes in the area of mutual understanding between people and system, mediated by digital artificial systems, as follows:
\begin{itemize}
    \item \textbf{Sharing knowledge} (corresponding to \emph{shifting perspective}) defined as the ability of the involved agents to share a common conceptual model about the domain of interest (e.g., an ontology to represent knowledge in a given domain);
    \item \textbf{Exchanging knowledge} (corresponding to \emph{managing transactions}) defined as the ability to meaningfully exchange information between two or more agents;
    \item \textbf{Governing knowledge} (corresponding to \emph{establishing rapport}) defined as the set of processes, roles and rules to regulate the sharing and exchange of knowledge between the involved actors.
\end{itemize}

\noindent In this chapter, we intend to analyse various scenarios of mutual understanding between people and systems supported by Neurosymbolic AI, 
highlighting how they potentially require solving issues in the three key processes of sharing, exchanging and governing knowledge.

Finally, we will characterise each described scenario, by identifying the involved agents (both human users and artificial systems, whether they are based on symbolic, sub-symbolic and neurosymbolic AI or on any other technology), the role they play in the scenario and the sources of knowledge they access to (e.g. data, schema, documents, machine learning models or even tacit knowledge).


%% file: sezioni/coney-gwap.tex
In this section, we address the problem of easing the collection, interpretation and analysis of structured data from people through dedicated applications and questionnaires, in order to facilitate automation of data analysis and reuse as well as longitudinal and comparative studies, by enabling \emph{mutual understanding between researchers who want to exchange and reuse data collected from human respondents}.

\subsection{Problem Scenario}
Jaime launched a citizen science project to involve people around all over the world to collect data about environmental pollution. He employed a gamified application to engage people to classify pictures that can be used to improve the understanding of the pollution issue. Through a dedicated campaign, he managed to engage a few hundred people in all five continents, who helped classifying some tens of thousands pictures; however, he noticed a varying level of quality and participation between the contributors, with some of them very active and others who seemed to lose interest over time.

Jaime wants to understand if it is worth continuing involving people to perform the classification task -- which indirectly contributes to raising awareness about environmental challenges -- or if he should use the classified pictures as a training set for a machine learning classifier. Moreover, he wants to understand what are the factors that influence their motivation to participate to his citizen science initiative and decides to run a survey, asking all the participants in his network to compile a structured questionnaire. 

Since he is a scientist, he performs a comparative study to contrast people and machine classification and he adopts questionnaire design best practices and reuse a motivation analysis framework from literature, because he want to compare his findings with the previous works. However, he knows from experience that, once he collects people responses, he will end up doing a lot of data cleaning and preparation before running his analyses. Jaime would love to find a solution to simplify the design, collection and analysis of his questionnaire data, also enabling other citizen science researchers to potentially repeat his investigation and compare their findings.

\subsection{Technical Solution}
\subsubsection*{Games with a purpose to engage people in collecting data}
Human Computation~\cite{law2011human} is the research area devoted to harnessing human intelligence to solve computational problems that machines find hard to solve; Games with a Purpose (or GWAP)~\cite{vonahn2006games} are a Human Computation approach that, in order to solicit people participation in solving computational problems, adopts gaming applications to provide an intrinsic incentive.

In the context of the collection and sense-making of structured data (linked data, knowledge graphs), GWAPs proved to be a suitable solution to engage people and we built a framework~\cite{re2018framework} to ease the setup of gamified applications for data linking, specifically for link creation, link ranking and link validation; moreover, we showed how to dynamically compute the ``mutual understanding" between human contributors with an incremental truth inference algorithm~\cite{celino2020refining}.

\begin{figure}[thb] 
	\centering
	\subfloat[Classify an image]{\label{fig:game}\includegraphics[width=.305\textwidth]{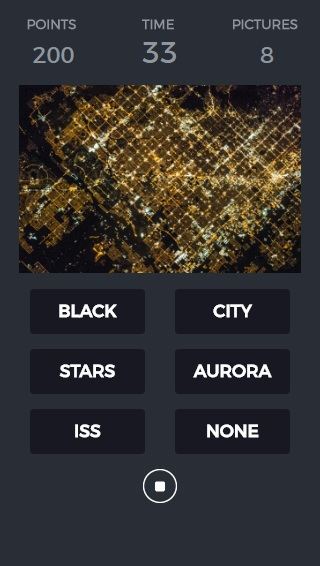}} 
	\quad
	\subfloat[Agreement]{\label{fig:agreement}\includegraphics[width=.300\textwidth]{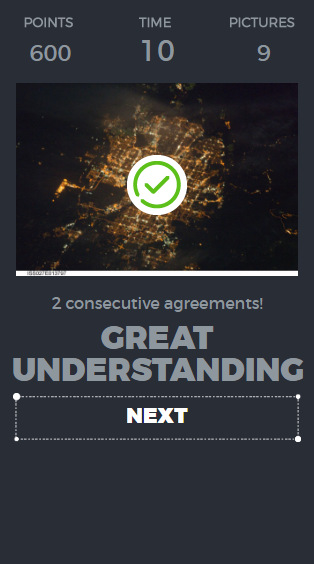}} 
	\quad
	\subfloat[Disagreement]{\label{fig:disagreement}\includegraphics[width=.303\textwidth]{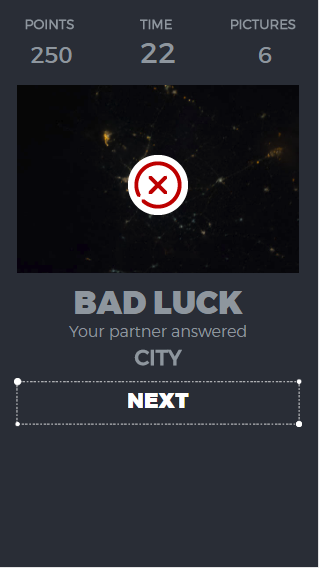}} 
	\caption{Game with a Purpose for data linking and image classification}
	\label{fig:gwap}
\end{figure}

We adopted the GWAP approach to build Night Knights~\cite{re2018human}, a gamified application to classify pictures in the area of light pollution (cf. also Figure~\ref{fig:gwap}) and we managed to involve $\sim$1.000 players to classify $\sim$40.000 pictures over a couple of months.

\subsubsection*{Patterns in human and machine learning involvement for classification}
\begin{figure}[htb]
	\centering
	\includegraphics[width=.8\columnwidth]{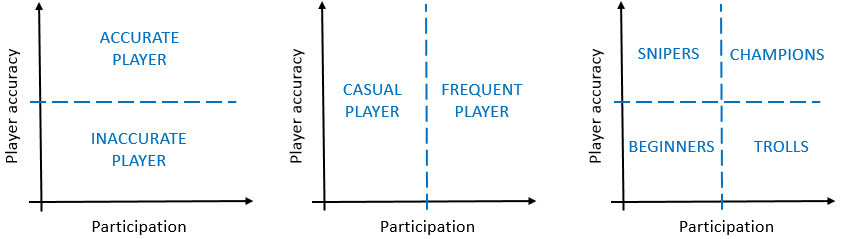}
    \\~\\
	\includegraphics[width=.7\columnwidth]{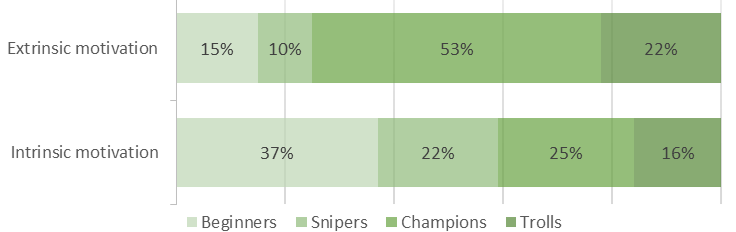}
  	\caption{Definition of GWAP player profiles and their distribution with different incentive mechanisms}
	\label{fig:gwap-profiles}
\end{figure}

When involving people in human computation effort to solve computational problem, the analysis of their behaviour reveals several interesting characteristics to investigate their role and the relevance of their contributions. We studied the participation data of the already mentioned Night Knights game to uncover the relationship between game incentives, player profiles and task difficulty~\cite{re2018interplay}; some example findings are displayed in Figure~\ref{fig:gwap-profiles}.

\begin{figure}[htb]
	\centering
		\includegraphics[width=1\columnwidth]{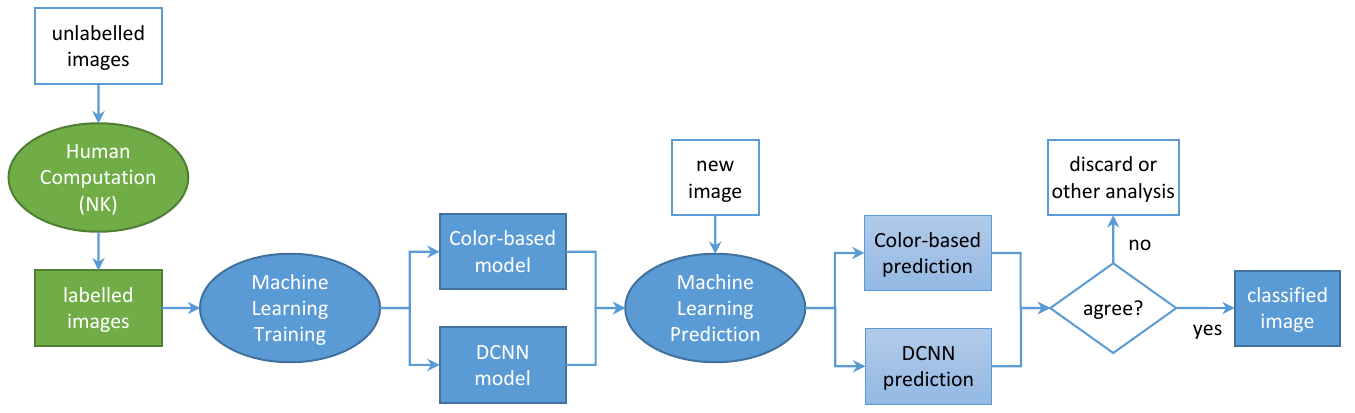}
	\caption{A possible combination of human and machine processing in a hybrid classification solution}
	\label{fig:hcomp-process}
\end{figure}

Such analysis also serves to the purpose of understanding how to build hybrid human-machine data processing pipelines, to make the best out of both worlds~\cite{re2018human}. Is it better to rely on human capabilities, asking people to perform tasks, or to train a machine learning system to automatically solve the problem? The answer largely depends on the specific case and the required accuracy: humans may be more reliable -- especially if they are domain experts -- but automatic processing can be cheaper. We performed an experimental comparison of different Human Computation and Machine Learning approaches to solve the same image classification task: we compared and contrasted them in order to come up with a long term combined strategy to address the specific issue at scale. While it is hard to ensure a long-term engagement of users to exclusively rely on Human Computation, manual classification is indispensable to overcome the ``cold start'' problem of automated data modelling (cf. hybrid human-machine workflow in Figure~\ref{fig:hcomp-process}).

\subsubsection*{Survey research as research objects}
Another way to study people participation to collaborative efforts is to directly ask them about their subjective perception through a quantitative research survey. That is what we did to analyse the factors that influence active participation to citizen science and crowdsourcing~\cite{celino2021participant}. In order to do so, we designed a survey to investigate different factors, we implemented it in a conversational survey tool~\cite{celino2020coney} that allows for semantically tagging of questions and answers and for collecting structured data described with respect to a survey ontology~\cite{scrocca2021survey} out of it.

The Survey Ontology\footnote{Cf. \url{https://w3id.org/survey-ontology}.} satisfies different requirements that often emerge when designing a questionnaire to collect quantitative data: (1) making the survey structure available as structured data, to make its knowledge explicit; (2) annotating questions with the respective investigated variables and (3) annotating answers with their numerical coding, to ease data analysis; (4) making the collected answers available as structure data, to enable comparisons and longitudinal studies; (5) keeping provenance of answers, to trace back in case of need; and (6) sharing the survey methodology, to foster repeatability and reproducibility of research. The ontology reuses concepts and relations from PROV-O  and  the Research Object suite of ontologies~\cite{belhajjame2014research}: this allows to package a composite artifact that represents, annotates, and shares a representation of both the questionnaire structure and the gathered survey responses.

\begin{figure}[htb]
  \centering
  \includegraphics[width=.8\columnwidth]{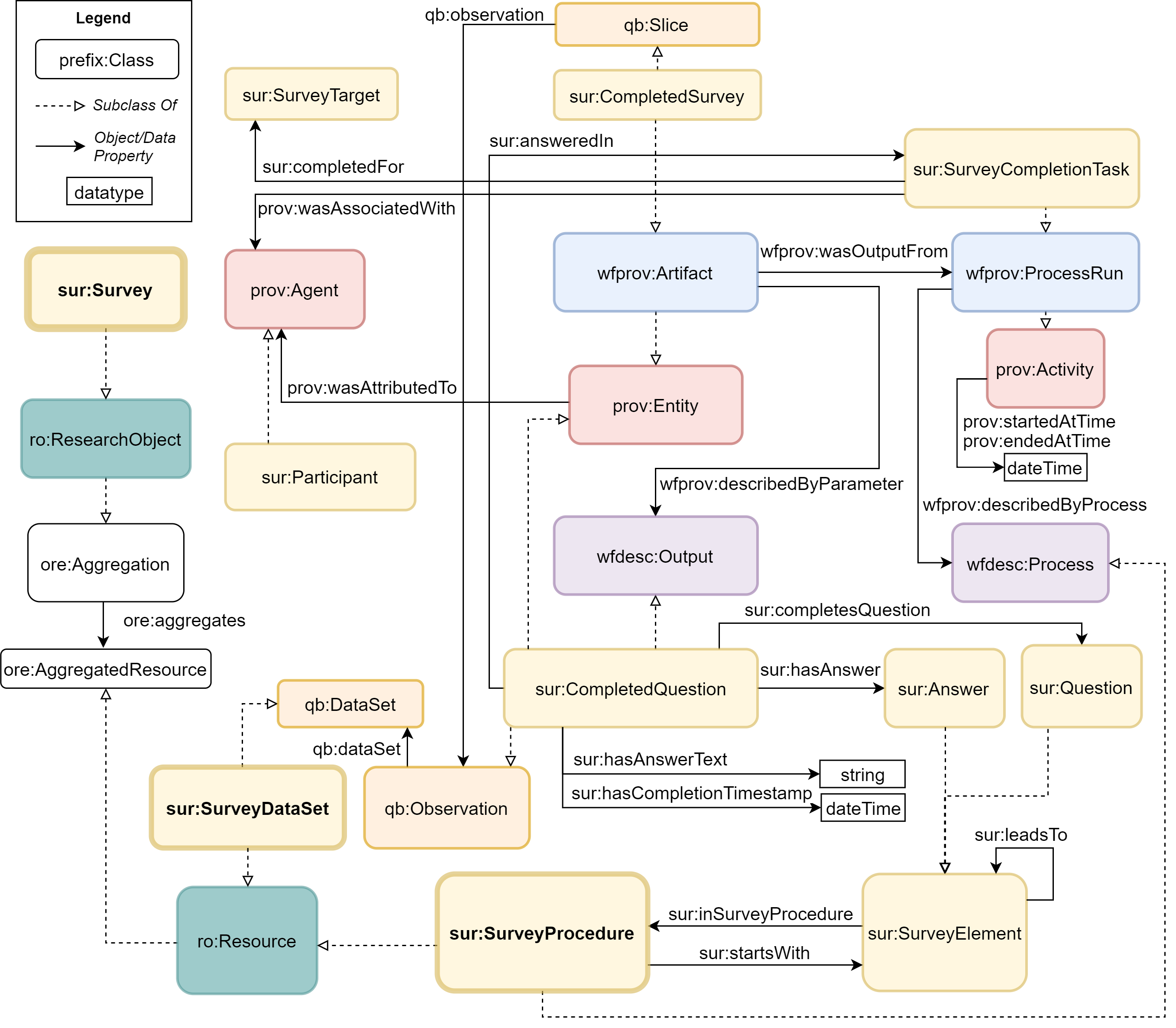}
  \caption{Overview of the Survey Ontology}
    \label{fig:extended}
\end{figure}

\subsection{Positioning w.r.t. Mutual Understanding Framework}
The main source of knowledge in this scenario is constituted by \emph{people}, because they provide their expertise or their ability to recuperate important information. People's knowledge is then collected, saved and managed in various \emph{information systems} (e.g. the GWAP application, the survey platform, the data analysis tool), which may support the long-term preservation of such knowledge.

The main mutual understanding problem addressed in this section is related to \emph{sharing knowledge}: involving people in a data collection and processing pipeline through a Human Computation approach means that we leverage their ability to provide their expertise and knowledge in solving a computational task, paving also the way to a hybrid human-machine processing (potentially also human-neurosymbolic); moreover, the goal of adopting a survey conceptual model to describe a questionnaire, a knowledge graph to collect respondent data and a research object to pack together methodology, data and analysis, is to facilitate and standardize the reuse of investigation approaches across different studies. The research object artifact also supports the \emph{governing knowledge} challenge, in that it provides a way to keep track of the entire survey campaign and to make its analyses reproducible, and partially also the \emph{exchanging knowledge} challenge, in that it allows the exchange of survey methodology and results between researchers, to perform comparative or longitudinal studies.

\subsection{Open Challenges and Possible Extensions}
Combining symbolic and sub-symbolic approaches to build composite patterns of data processing is the focus of the well-known "boxology" of design patterns for hybrid learning and reasoning systems~\cite{vanharmelen2019boxology}, as well as the topic of the analysis of mixed machine learning-semantic web systems~\cite{breit2023combining}. Further extending those works to take into account also the \emph{human component} would lead to the definition of hybrid human-neurosymbolic design patterns for intelligent systems that bring together artificial and human intelligence; in this respect, the recent wave of Generative AI solutions, with their human-like capabilities, seems to suggest that this combination will lead to a \emph{new generation of hybrid systems} to solve complex challenges in a more effective way.

With respect to questionnaires to collect and analyse quantitative data, our conversational survey tool~\cite{celino2020coney} was designed to provide respondents with a chat-like interactive experience to improve their engagement. The current and constant improvements to Conversational AI systems due to the advent of large language models are very likely to bring new forms of interactive survey systems. While in literature several works highlight the ability of chatbots to administer interviews and to increase participant engagement~\cite{xiao2019tell}, trust~\cite{akbar2018effects} and self-disclosure~\cite{zhou2019trusting}, a significant improvement could come from successfully mixing those neural approaches with symbolic ones to achieve structured data collection and answers' classification and coding.

%% file: sezioni/khub-perks.tex
In this section, we address the problem of extracting knowledge from unstructured data sources and of making tacit knowledge explicit by involving people in knowledge acquisition by enabling \emph{mutual understanding between people who need to transfer/acquire knowledge, possibly supported by digital systems}.

\subsection{Problem Scenario}
Jenny is a product manager of a manufacturing company. She is responsible to ensure that the supporting documentation for the use, repair and troubleshooting of her products is available, clear and up-to-date. Her team always makes an effort to build manuals containing all the relevant information. 

However, Jenny realises that, even if her team always makes an effort to build manuals containing all the relevant information, two problems often arise: (1) this information is not easily accessible, as manuals are often very long; for example, a maintainer who is asked to repair a product upon customer request may need to find the right document and, within the document, to find the relevant information among a lot of other less useful detail for the problem at hand; (2) a part of the knowledge remains tacit in the head of people; for example, a novice or less experienced employee may often need  to ask for advice to senior experts, because not all the relevant information can always be documented, some knowledge remains implicit or is acquired only through direct experience. Jenny would love to find a solution to extract relevant knowledge from documents and to facilitate the explicitation of the tacit knowledge of experienced colleagues.

\subsection{Technical Solution}

\subsubsection*{Information retrieval and structured knowledge} 
The desire to find the relevant information for the task at hand is a common challenge in the industrial sector~\cite{rula2022eliciting}. Finding the right piece of knowledge or the right answer is the main basis and motivation for any information retrieval and question answering approach. More recently, the advent of Generative AI led to the rise of Retrieval-Augmented Generation approaches (RAG~\cite{lewis2020retrieval}), which combine vector indexes with pre-trained parametric models. Among different RAG approaches, Knowledge Graph-based RAG architectures are emerging, exploiting the power of structured information -- also described with respect to an ontological conceptual model -- together with neural approaches.

\begin{figure}[htb]
    \centering
    \includegraphics[width=0.9\textwidth]{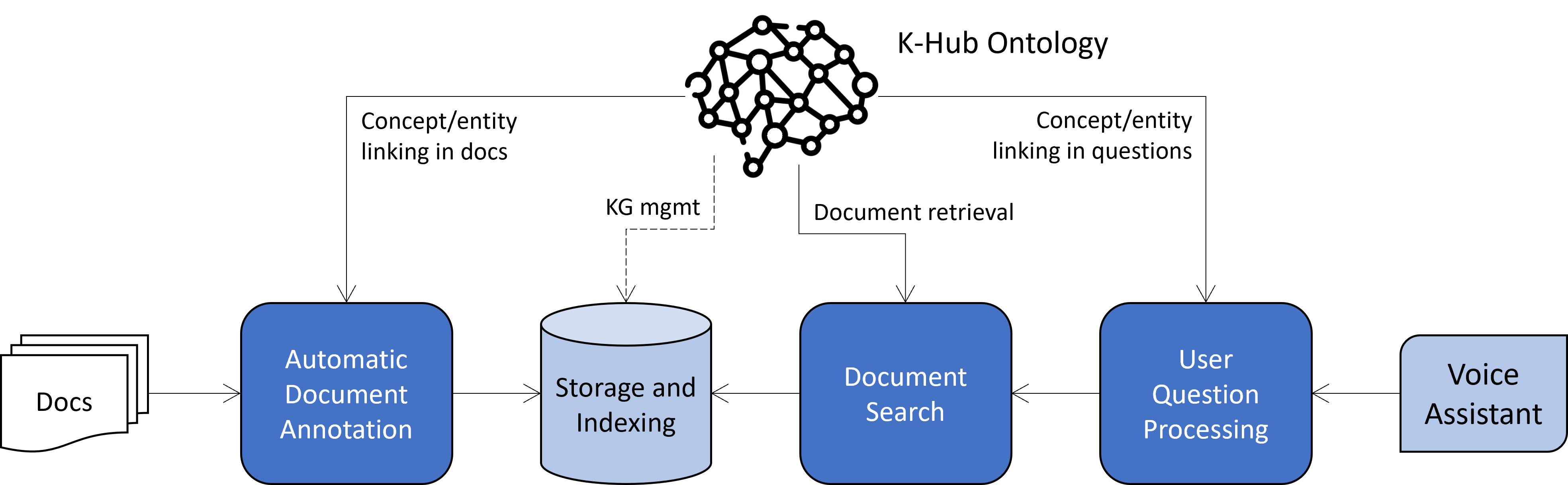}
    \caption{Use of the K-Hub ontology~\cite{rula2023khubonto} in a document processing and search pipeline}
    \label{fig:k-hub-onto-use}
\end{figure}

It is indeed well-known that structured data makes it easier to find relevant information, and that knowledge graphs are a powerful solution to model, share, and preserve structured descriptions of relevant entities and relationships. For this reason, we have been working on building a modular ontology to facilitate document retrieval and knowledge extraction in industrial settings~\cite{rula2023khubonto}. The proposed ontology can be used at different stages of annotation, storage/indexing, search and query answering (cf. Figure~\ref{fig:k-hub-onto-use}).

\subsubsection*{Knowledge extraction from unstructured information}
Traditional and emerging NLP techniques can be used to extract relevant information form text and other kinds of unstructured sources: when knowledge is embedded in documents (but also pictures and other multimedia formats), it is common to perform (named) entity recognition, relation extraction and entity linking. The goal is to turn unstructured information into structured descriptions, often in the form of a knowledge graph, according to a predefined ontology and possibly linking to pre-existing knowledge bases.

In the area of process and procedural knowledge, some work exists to automatically or semi-automatically extract relevant information from textual descriptions, also leveraging LLM capabilities~\cite{bellan2022extracting,rula2023procedural}. Still, a holistic solution to this challenge is still missing, as documents may be incomplete (e.g., information given for granted, background knowledge, common sense, contextual data) or they may require specialised expertise to be interpreted and understood. Furthermore, the same piece of information sometimes can be made explicit in similar but different ways, making it difficult to completely automate the knowledge extraction task~\cite{carriero2024human}.

\subsubsection*{Knowledge acquisition from people}
In order to correctly extract relevant information from unstructured sources, it is of course also possible to ask knowledgeable people to manually annotate documents, especially when the user is guided in identifying and tagging with an explicit semantics~\cite{rula2023annotation}; however, this kind of manual task is known to be a tedious and expensive work, and also very error-prone.

Moreover, when knowledge is not documented at all, because it remains tacit, i.e. ``in the head" of the human expert, no kind of manual or automatic annotation can be adopted. In this case, the need to make tacit knowledge explicit calls for acquisition solutions that guide the user to collect and structure their specialised information, in order to facilitate transfer and preservation of such knowledge. Whether knowledge is (semi-)automatically extracted from documents or acquired from people, it is very common to adopt \emph{human-in-the-loop} approaches for knowledge validation, as in \cite{tsaneva2024enhancing}: building hybrid information processing pipelines that combine human and machine intelligence can help finding the best trade-off between quality and cost.

\subsection{Positioning w.r.t. Mutual Understanding Framework}
The main sources of knowledge in this scenario are \emph{unstructured information} (especially documents) and again \emph{people} (for tacit knowledge); the challenge is to populate and maintain up-to-date structured sources for long-term preservation and access.

The main mutual understanding problem addressed in this section is related to \emph{sharing knowledge}, in that the goal of collecting information from documents or people is to facilitate its later reuse by other (artificial or human) users. When made explicit and structured, the elicited information could also support \emph{exchanging knowledge} between different actors. Finally, a holistic neurosymbolic solution to the extraction/acquisition challenge would also need to address the \emph{governing knowledge} dimension, in that it would help in the long-term management of domain knowledge.

\subsection{Open Challenges and Possible Extensions}
Regarding the knowledge extraction from text, there are currently different generative AI-based approaches to support tasks like entity recognition and relation extraction~\cite{kumar2020,ding2021few,shi2024generative}; however, the precision and reliability of such extraction is still far from ideal, especially when the documents does not contain the entire contextual knowledge, because it is implicit or common sense.

Regarding the knowledge acquisition from people, as discussed, involving domain expert is a very expensive task and it is not even always possible. To this end, neurosymbolic approaches for semi-automatic annotation/collection with human-in-the-loop seem to be a suitable research direction, especially adopting mixed and iterative strategies: people could be supported or guided in the process, either by leveraging a conceptual domain model (e.g. to give the expert a ``structure" to follow to make their knowledge explicit) or by providing an interactive solution to ease the acquisition (e.g. using conversational AI to facilitate knowledge disclosure~\cite{zhou2019trusting}).

%% file: sezioni/sprint-tangent.tex
In this section, we address the problem of enabling data interoperability among information systems. Interoperability means reconciling heterogeneous data sources, so to facilitate integration and reuse, i.e. to enable \emph{mutual understanding between different information systems that need to meaningfully exchange information}.

\subsection{Problem Scenario}

Tiago works for a company that realises digital solutions for traffic management in urban environments. In particular, his team develops a decision support system (DSS) for city mobility officers that includes near real-time and forecast information about city traffic. This system gathers data from multiple information sources (e.g., public transport vehicle tracking, traffic lights configuration, video cameras monitoring car flows, etc.), processes incoming data through machine learning solutions, and displays the current and forecast status on an interactive dashboard.

Every time a new city approaches Tiago to customize the DSS solution, he faces significant challenges. Despite the existence of (open) data platforms collecting potential input data sources, Tiago often has to deal with disparate metadata schemas and custom APIs, which further complicate data integration. He needs to identify and understand every available information source, decode their often undocumented data formats and conventions, and perform extensive data pre-processing and transformation to integrate this data into the DSS.

Additionally, Tiago's team requires historical datasets to support the development of tailored machine-learning systems for the DSS. These datasets are often unavailable and should be collected from real-time data sources. Also in this case, harmonising data according to a common data format is crucial to facilitate the training of models considering different information sources from several cities.

These tasks are tedious and expensive, yet essential for ensuring compatibility between the information sources and the DSS solution. Tiago is seeking a way to simplify the process of achieving mutual understanding between the diverse information sources of each city and his DSS.

\subsection{Technical Solution}

\begin{figure}[htb]
	\centering
		\includegraphics[width=1\columnwidth]{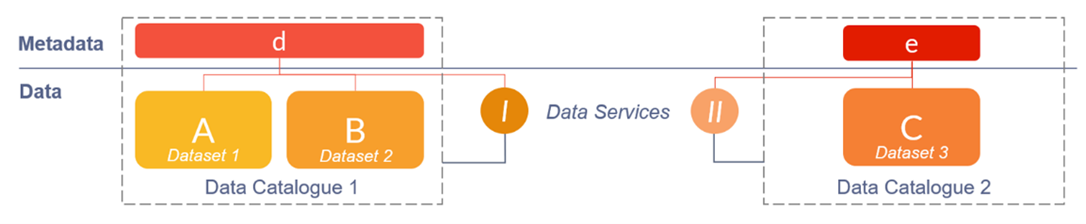}
	\caption{Generic scenario requiring interoperability across information systems.}
	\label{fig:problem-tangent}
\end{figure}

In the generic scenario depicted in Figure \ref{fig:problem-tangent}, open and private data catalogues describe and host data from different information systems. Each data catalogue may contain: (i) datasets in different formats and/or using different data models (A, B, C); (ii) data services relying on heterogeneous specifications and technologies (I, II); (iii) data sources described according to different metadata specifications (d,e).

The definition of a proper solution to this heterogeneity is not straightforward and should face five challenges (Locate, Access, Harmonise, Integrate, Extract).

Identifying a single solution to the abovementioned challenges is impossible since a single interoperability problem cannot be formulated \cite{sadeghi_interoperability_2024}. Indeed, the identified data interoperability issues are widely heterogeneous and pose various requirements that can be possibly faced only by considering a set of tools appropriately configured.

\begin{figure}[htb]
	\centering
		\includegraphics[width=1\columnwidth]{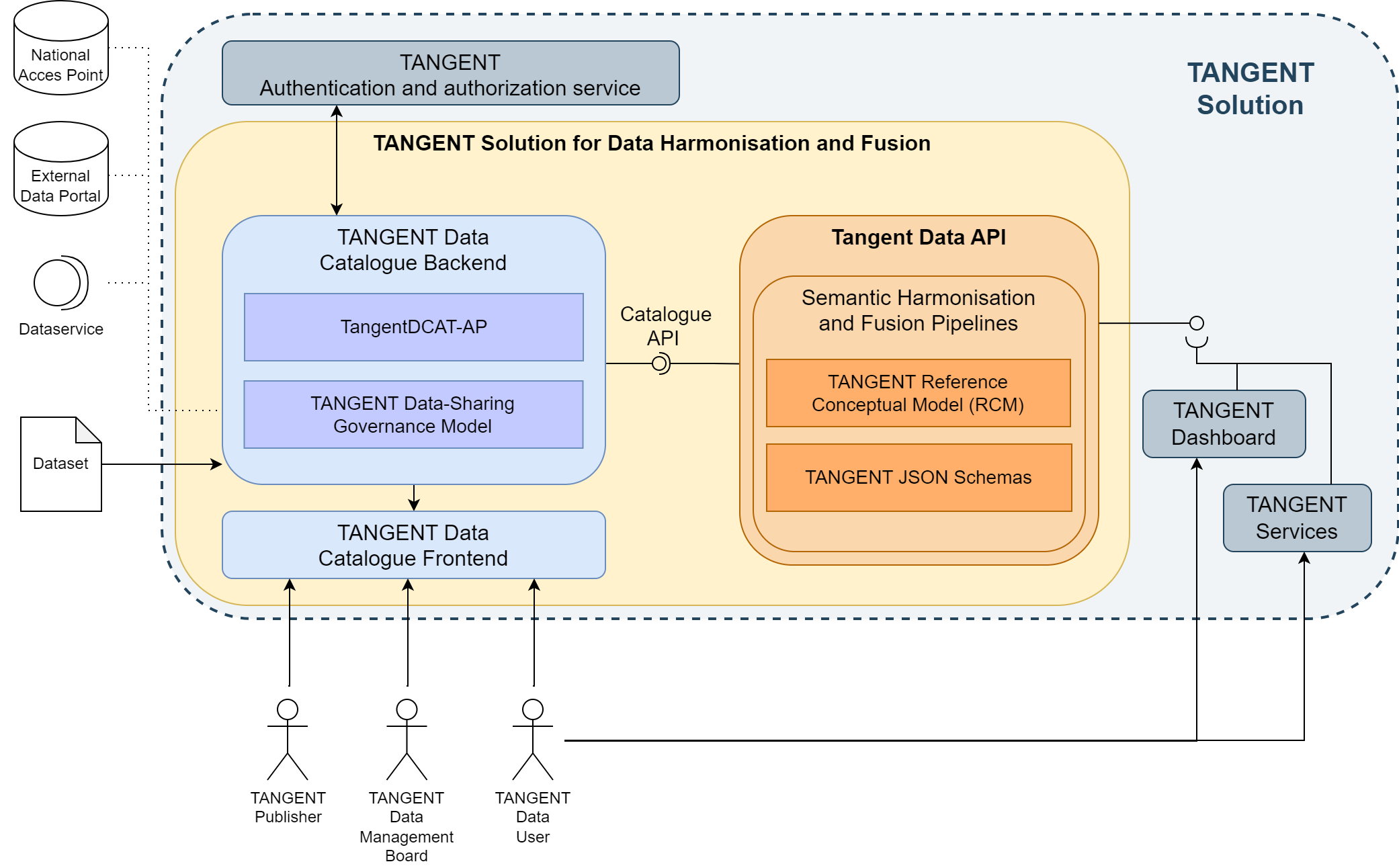}
	\caption{End-to-end solution for data interoperability in the TANGENT project.}
	\label{fig:tangent-simple}
\end{figure}

We developed an end-to-end solution for data interoperability based on Knowledge Graph technologies to solve the heterogeneity issue in traffic management highlighted above in four different European cities~\cite{scrocca2024intelligent}. The proposed solution, represented in Figure~\ref{fig:tangent-simple}, deals with the mentioned challenges and proposes an integrated set of components that can be configured and extended to address interoperability issues in a generic domain. 

The first challenge is the findability and discoverability of data. Data cannot be re-used and (made) interoperable if they cannot be found. For this reason, data catalogues/portals are implemented 
to describe data sources 
through a set of metadata. The challenge is associated with the need for a proper, structured and machine-readable description of data sources that could also support interoperability across different data catalogues. Once data sources are located, the second challenge is related to data accessibility. Data catalogues adopt different strategies for data access mainly associated with the architectural choices for the hosting and storing of static and dynamic data sources. The challenge is to enable uniform access to heterogeneous data sources by external information stystem.

For these reasons, the first two core components of the proposed solution are a shared \emph{Data Catalogue} to enable the findability of data sources and a uniform \emph{Data API} for accessibility. The \emph{Data Catalogue} provides a user interface and an API to discover all the relevant data sources harvested by other data portals and or directly described and published by users. Data interoperability is guaranteed by adopting a common metadata specification based on well-known vocabularies such as DCAT-AP \cite{DCAT-AP} that may be extended to support the specific requirements of a domain or customer scenario. Moreover, being the point of access for users interacting with data sources, the \emph{Data Catalogue} also enables the enforcement of processes and rules according to a common data-sharing governance model~\cite{tra2024governance}.
The \emph{Data API}  provides uniform access to data sources collected through the Data Catalogue and represents the integration point for the downstream information systems accessing the data sources.

The remaining three data interoperability challenges (harmonise,
integrate, and extract) are related to the processing of (meta)data to enable their integration and exploitation according to common semantics. A flexible solution is required to address heterogeneous requirements in terms of: 
(a) schema and data transformation: information manipulation to obtain syntactic (structural) and semantic
interoperability of (meta)data;
(b) integration with existing information systems as data sources (i.e., components generating or storing the data) and/or data sinks (i.e., components consuming or archiving data).

Different approaches can be exploited and implemented, spanning from ad-hoc solutions targeting a specific scenario to more general and scalable
solutions supporting multiple stakeholders and data representations. The semantic any-to-one mapping approach based on~\cite{Vetere2005Models} and validated in~\cite{Scrocca2020Turning}
reduces the number of mappings, i.e., translations from one
representation to another, that are needed to implement interoperability by
different stakeholders. Such an approach is based on identifying a reference model for the domain of interest. Each stakeholder is responsible for defining mappings from their own data representation to the reference model (\emph{lifting}) and vice versa (\emph{lowering}). Additionally, the harmonisation, integration and extraction process may require the definition of custom pre- and post-processing, considering different interoperability issues.

To address these aspects, two additional components are identified to support the solution: a \emph{Reference Conceptual Model} defining common semantics and the composition and configuration of \emph{Semantic Harmonisation and Fusion Pipelines}.

The \emph{Reference Conceptual Model} supports representing heterogeneous information from different data sources through a common ontological model to enable shared semantics and interoperability. The model is based on existing data standards to adopt the correct domain terminology and covers the representation of all the entities and properties required to implement meaningful data exchanges among the involved stakeholders.

A flexible and scalable technology for the implementation of the pipelines should provide (i) a set of reusable building blocks that can be configured according to specific requirements, and (ii) a declarative approach to configure the lifting and lowering transformations without developing ad-hoc and hard-to-maintain solutions.
Chimera\footnote{\url{https://github.com/cefriel/chimera}}
\cite{Grassi2023Composable} is an open-source solution based on Apache Camel to enable the definition of semantic data transformation pipelines with different components for knowledge graph construction, transformation, validation, and exploitation.
The advantage of Chimera is its integration with Apache Camel, providing off-the-shelf and production-ready components to implement Enterprise Integration Patterns and to integrate pipelines with heterogeneous systems (e.g., HTTP API, WebSocket, MQTT). 

The implemented pipelines are documented within the \emph{Data Catalogue} to publish the description of newly generated data sources or to describe the availability of a data source according to a specific format. The \emph{Data API} integrates the integration and execution of the pipelines to provide access to data sources according to the requested format.
Such an approach can be applied for the harmonisation of heterogeneous metadata to enable the federation of existing data portals \cite{carenini_enabling_2021} or directly for the integration and fusion of data sources \cite{tra2024harmonisation}. Moreover, the application of pipelines over time to collect, harmonise and store data can be leveraged to generate historical datasets for the training of machine learning models.

\subsection{Positioning w.r.t. Mutual Understanding Framework}
The main source of knowledge associated with this scenario is related to the structured \emph{metadata and data managed by information systems}; the challenge is to enable easy discoverability, access and exchange of such knowledge.
The problem presented in this section is mainly related to the \emph{exchanging knowledge} challenge, as the mutual understanding issue is mainly related to the possibility of enabling a meaningful exchange of data between the information systems. However, the adoption of a shared conceptual model also addresses the \emph{sharing knowledge} dimension while the catalogue, as well as the API layer, are meant to support the \emph{governing knowledge} aspect.

\subsection{Open Challenges and Possible Extensions}
The semantic interoperability solution described above contributes to the mutual understanding between the systems that need to meaningfully exchange information by means of two important types of artifacts: (1) data and metadata models, expressed as ontologies, vocabularies and metadata profiles, and (2) mappings that encode the correspondences between the conceptual model and the data formats adopted by the different sources and systems.
Considering the first artifact, while the proposed technical solution mainly leverage symbolic approaches, the interoperability of data and metadata is essential to support the retrieval and exploitation of data sources for machine learning~\cite{mubashara2024}.

With respect to the latter, several approaches exist to define those mappings. Declarative approaches for knowledge graph construction can effectively support lifting transformations to RDF, while a standardised lowering solution to convert RDF to any format using a generic declarative language is currently missing~\cite{scrocca2024not}. 
Despite the encoding and execution of the mapping, a well-known challenge is related to the identification of the conceptual mappings, which is a very expensive and intensive activity. Some recent approaches aim at adopting neurosymbolic approaches, by exploiting language models to identify similarities among concepts and generate the mappings \cite{randles_r2rml-chatgpt_2024,hosseini_automated_2019,hofer_towards_2024}. While a complete automation of the mapping generation is not yet achievable, the current ability of Generative AI solution seems promising to support a domain expert in the semi-automatic definition of mappings.
Future research directions may also investigate the potentiality of leveraging non-symbolic models to transform data directly from one format to another. A multi-agent approach may support data harmonisation through the cooperation of not-symbolic transformations and iterative corrections based on the symbolic validation of results according to the target data schema (e.g., ontology). 

On the other side, open challenges are also related to the configuration of pipelines for data exchanges. Low-code and no-code solutions are proposed to facilitate these tasks and the integration with code generation techniques from natural language can further facilitate the user in this task \cite{nikitha2024}.

%% file: sezioni/daydreams-ride2rail.tex
In this section, we address the problem of guaranteeing reproducibility and trustability in a digital environment involving different stakeholders, i.e., to enable \emph{mutual understanding of operations executed by users with different roles and responsibilities through digital artefacts}.

\subsection{Problem Scenario}

We present two scenarios where multiple stakeholders should cooperate and act together within a digital environment. The first scenario focuses on monitoring the deployment, execution and use of intelligent asset management systems relying on AI-based approaches for prescriptive maintenance. This scenario involves the integration and reuse of components developed by different stakeholders requiring \emph{mutual understanding between people and data pipelines}.
The second scenario addresses the challenge of defining, applying and exchanging business agreements between different actors that should cooperate and trust each other. The objective is to enable \emph{mutual understanding of the terms and conditions of a business agreement} between the stakeholders involved.

\subsubsection*{Intelligent asset management system (IAMS)}
Renzo is an infrastructure manager, responsible for the maintenance of a complex railway network. To ensure the proper operations of the railway, Renzo coordinates a group of different stakeholders with challenges about data management, runtime interactions and trust among the parties. They have different digital systems, including predictive and prescriptive machine learning models that are used to manage the maintenance activities, by foreseeing the need for an intervention on the network (e.g. the need for a future repair work on the trails) and by scheduling the actual work of the operators (e.g. the planning of employees turns).

If something does not go as planned (e.g., failure of the network due to maintenance issue), it is of utmost importance for Renzo to audit the operations executed by people and systems, to safely reproduce what happened to identify the root cause, and to inject the new knowledge in the system so to ensure that future planning avoids repeating mistakes and learn for the past experience. Renzo would love to find a solution to reliably execute and monitor machine learning pipelines and to create a trusted environment for governing and coordinating the different systems and people, thus enabling mutual understanding between all the involved stakeholders.

\subsubsection*{Business agreements for multimodal transportation}
The new ride-sharing service RideWithMe would like to provide drivers and passengers with a trusted service to organize shared rides and to cooperate with other transport operators to promote multimodal journeys.
A shared ride booked via RideWithMe can be seen as a business agreement between the user offering a shared ride with her/his car and the passengers. Each driver can be considered a private transport service provider (TSP) offering transportation services on a specific route. The passenger accepts to pay a certain price in exchange for a shared ride from a certain location to a specific destination. Such information should be shared in a trusted way so that RideWithMe can ensure dispute resolution mechanisms among the driver and the passengers. RideWithMe leverages the blockchain and smart contracts to guarantee trust among the parties.
Similarly, RideWithMe and other transport service providers agreed to use smart contracts to support the definition of multimodal packages for travellers. 
In both cases, even if a shared data model ensures interoperability in data exchanges among information systems, the implementation-dependent nature of smart contracts makes it difficult to represent univocally the terms and conditions of the agreements using the proper domain terminology. Indeed, while the adoption of blockchain technology ensures trustability, it affects the mutual understanding of the implemented business agreements.

\subsection{Technical Solution}

The technical solutions proposed leverage Knowledge Graphs to support the reproducibility of data-driven solutions and trustability over blockchain architectures in the multi-stakeholder environment.

\subsubsection*{IAMS integration support framework}

The adoption of intelligent maintenance solutions requires collecting data from different sources, implementing specific software components for data analytics and visualisation, and integrating them to support data pipelines. A significant challenge lies in the variety of stakeholders involved in deploying such solutions and executing the target maintenance processes.

To support these challenges, we proposed the IAMS Integration Support Framework~\cite{scrocca2024integration} 
based on three macro-functionalities: (i) a shared catalogue to enhance the findability and reusability of digital artifacts (IAMS Shared Catalogue); (ii) a runtime area for the integrated deployment and execution of artifacts associated with intelligent maintenance scenarios (IAMS Runtime Area); and (iii) a blockchain-based system to handle the monitoring and tracking of artifacts, processes and operations performed by users (IAMS Process Tracking).

\begin{figure}[!ht]
\centering
\includegraphics[width=.9\textwidth]{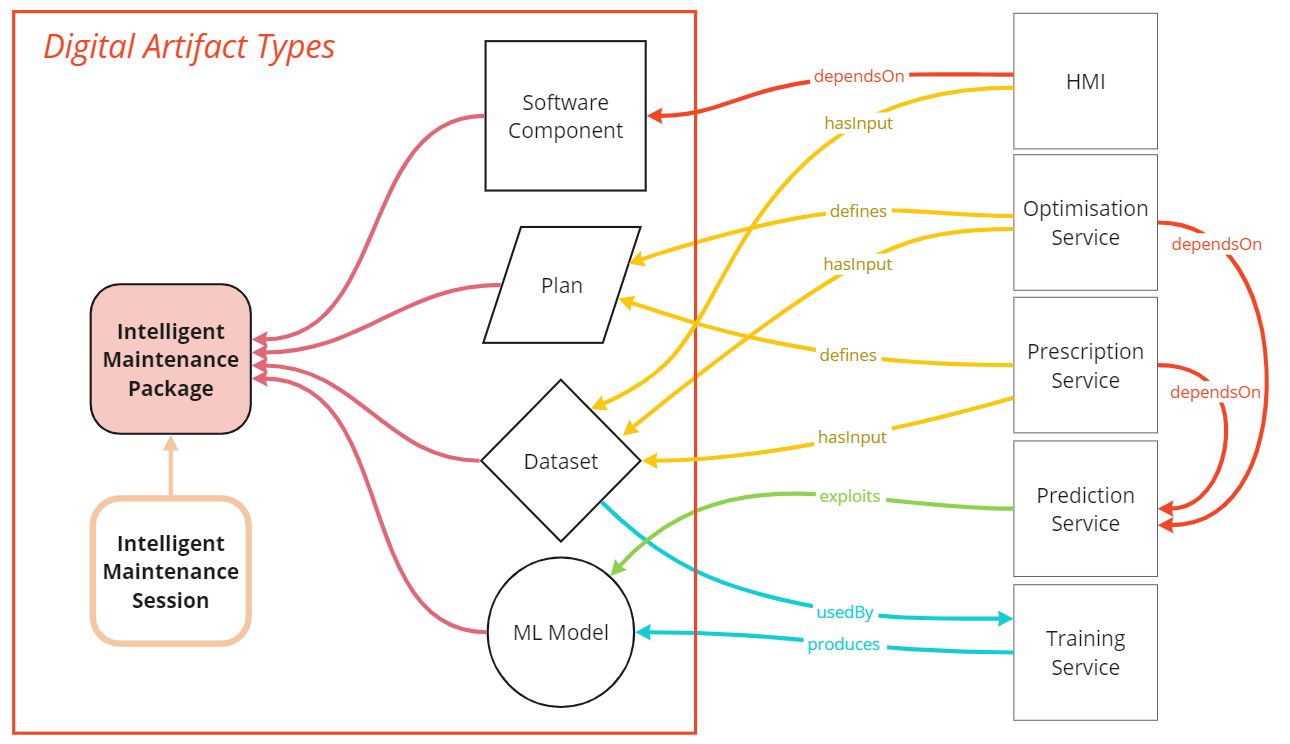}
\caption{IAMS Shared Catalogue: artifact types and their dependencies.}
\label{fig:asset-types-d52}
\end{figure}

Figure \ref{fig:asset-types-d52} provides an overview of the main components for an intelligent maintenance solution and their dependencies. We identified four basic digital artifact types: Dataset, Software Components, Machine Learning Model, and Maintenance Plan. For each of them, we defined a metadata schema relying on well-known vocabularies. 

Under the overall maintenance scenario, describing and tracking each digital artifact individually is not enough to foster reproducibility. For this reason, two additional digital artifact types are defined to fully describe the dependencies and usage of digital artifacts:
\begin{itemize}
    \item an \textbf{Intelligent maintenance package} (IAM Package) describes a maintenance scenario that leverages different digital artifacts and can be executed in an integrated runtime environment (e.g., the datasets supporting the training of a certain machine learning model, the training software component and the predict/prescribe components exploiting such model);
    \item an \textbf{Intelligent maintenance session} (IAM Session) describes the information about operations made by a user in the interaction with digital artifacts composing an \emph{Intelligent maintenance package} (e.g., a certain user requested the re-training of a machine learning model with different hyper-parameters and then generated a maintenance plan relying on the trained model).
\end{itemize}

A \emph{package} resembles the concept of a Research Object, mentioned in Section \ref{sec:coney-gwap}. It contains static information on \emph{digital artifacts} used for a particular scenario and the configuration for their integration. A \emph{session} contains provenance information obtained at runtime on the usage of a \emph{package}, e.g. the software component used to generate a machine learning model, and traced on the blockchain.

The \emph{RO-Crate} specifications can support the serialization of such information. Each digital artifact 
can be seen as a resource in a \emph{RO-Crate} \cite{rocrate} using a specific set of metadata. 

\begin{figure}[!ht]
\centering
\includegraphics[width=0.9\linewidth]{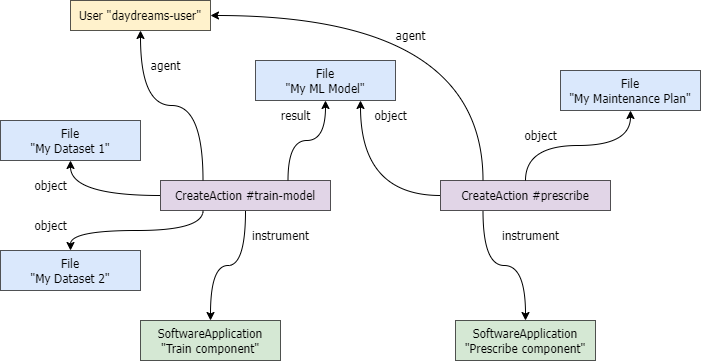}
\caption{Example of Process RO-Crate in the context of an intelligent maintenance scenario in the DAYDREAMS project.}
\label{fig:ro-crate-workflow}
\end{figure}

Considering an intelligent maintenance session, the \emph{Process RO-Crate} profile \cite{process-run-crate} can be used to represent all the lifecycle operations performed on the digital artifacts of the package, and all the runtime operations executed within an IAM Session. Figure~\ref{fig:ro-crate-workflow} represents an example Process RO-Crate describing two operations (\emph{CreateAction} entities in purple), their connection with specific digital artifacts (\emph{File} enties in blue, \emph{SoftwareApplication} entities in green) and the identification of the user (\emph{User} entity in yellow) that performed them. 
A \emph{train-model} operation is described by identifying the user requesting the training, the datasets used, the software component executed and the produced machine learning model. Similarly, a \emph{prescribe} operation can be described by identifying the user requesting the prescription process, the machine learning model used, the software component executed and the produced maintenance plan

The integration of such information as digital artifacts in the shared catalogue supports reproducibility and trustability for intelligent maintenance scenarios implemented through the \emph{IAMS Integration Support Framework}.

\subsubsection*{Ontological smart contracts for multimodal transportation}

In an ecosystem with various stakeholders, the implementation of business agreements through a distributed ledger provides several benefits regarding information trust and the automatic execution of the agreed terms modelled as smart contracts. This approach, however, doesn't provide any guarantee about the interoperability of the defined agreements from a technological and semantic perspective. On one hand, the domain terminology shared among the involved stakeholders should be referenced by the modelled entities, on the other hand, other software systems can benefit from a machine-readable representation of the agreements.

We proposed a solution to describe smart contracts implemented on the blockchain using an ontology \cite{ride2rail}.
The concept of \emph{Ontological Smart Contract}~\cite{cantone_ontological_2021} can be leveraged to model the semantic of business agreements.  
The following steps should be performed: (i) investigation of the business agreements to be modelled in the considered scenario (use cases and user stories), (ii) analysis of the domain terminology covered by the business agreements (facts and competency questions), (iii) identification of existing vocabularies covering the relevant domain entities and relationships, and/or implementation of an ontology supporting their representation, (iv) modelling of each business agreement as an ontological smart contract identifying the involved entities and the terms of the agreement, and, optionally, (v) representation of specific instances of the business agreement stored on the ledger using the ontology. In this way, different stakeholders are able to access through uniform terminology a description of the smart contracts and, possibly, their instances.

\begin{figure}[!ht]
\centering
\includegraphics[width=1\linewidth]{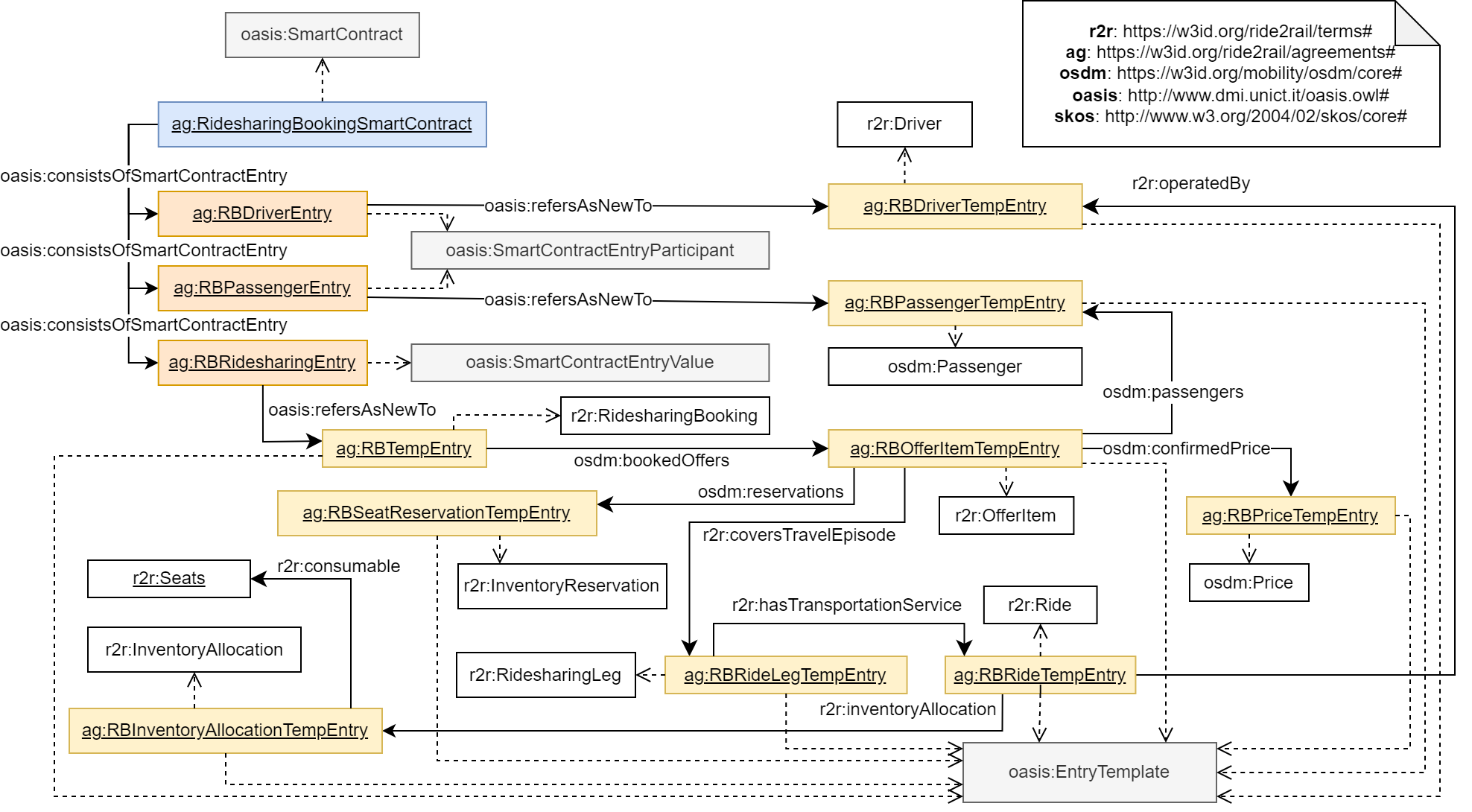}
\caption{Diagram describing the \emph{RidesharingBookingSmartContract} agreement}
\label{fig:ride2rail-agreement}
\end{figure}

The \emph{Ride2Rail Ontology for Agreements} (\texttt{https://w3id.org/ride2rail/terms\#}) enables the application of this concept to support the interoperability of business agreements in the multimodal transportation scenario detailed above. The ride-sharing booking is modelled as an agreement between the driver and the passengers for dispute resolution, and the definition of multimodal packages as agreements between different transport service providers. 
The RDF dataset with the modelled agreements is published at \texttt{https://w3id.org/ride2rail/agreements\#} and the agreement for a ridesharing booking is reported in Figure \ref{fig:ride2rail-agreement}.

This approach addresses two needs of mutual understanding: (i) the description in an implementation-independent way of the smart contracts defined according to a specific blockchain-based solution, and (ii) the adoption of proper and shared terminology to describe domain entities and their relationships. 

\subsection{Positioning w.r.t. Mutual Understanding Framework}
The main sources of knowledge in this scenario are \emph{provenance information} to guarantee reproducibility and auditing of operations performed by users and software components. The challenge is to trace and collect the relevant information while guaranteeing trust between the stakeholders.
The main mutual understanding problem addressed in this section is related to the \emph{governing knowledge} aspect, in that the final goal is to allow different stakeholders to correctly operate and to trust each other: the proposed solutions allow for monitoring and auditing the interconnected systems in a trustworthy way. The discussed IAMS solution highlights how symbolic approaches can support the governance of neural ones, enabling the reproducibility of pipelines and the auditing of performed operations.

In solving the governance aspect, the catalogue also provides an environment to share and access the relevant artefacts (datasets, software components, machine learning models, etc.) and their respective dependencies, also in accordance with the FAIR principles, thus directly addressing also the \emph{sharing knowledge} aspect. This is also enhanced by the adoption of the Research Object approach, to enable reproducibility and understanding of what happened in previous sessions. In solving the \emph{governing knowledge} aspect, the blockchain provides a mechanism for trust that should be complemented via a shared conceptual model (\emph{sharing knowledge}) to ensure mutual understanding among the involved parties.

\subsection{Open Challenges and Possible Extensions}

The main open challenge discussed in this section is related to the integration of neurosymbolic approaches not only as part of hybrid algorithms but also to support the sharing and governance of digital artefacts. The adoption of a structured metadata description can facilitate their findability and reuse, as discussed also in Section \ref{sec:sprint-tangent}, but should not be limited to isolated and static information. Additional metadata on how different digital artefacts are combined together and runtime information on their usage represents a key element for reproducibility. In this direction, additional information may be retrieved not only by tracing the interaction among the components but also by further instrumenting neuro-symbolic components for more granular auditing \cite{breit_combining_2023}.

An additional challenge is related to the difficulties of encoding and communicating all the relevant details of business agreements by adopting a fully symbolical approach. Indeed, often agreements leverage the blurred boundaries of legal terminology that can hardly be represented using a logical encoding. In this context, the adoption of LLMs can be leveraged to combine business agreement terms in structured and unstructured formats and provide natural language explanations and answers to the users involved.

%% file: sezioni/spatiotemporal-robots.tex
In this section, we address the problem of building reliable world models to support robots and autonomous agents in making sense of complex deployment environments. This model building process was originally introduced by Lake et al. \cite{lake_building_2017} as one of the essential requirements for achieving human-like intelligence in machines.  However, building and maintaining accurate world models is also crucial to achieve a \textit{mutual understanding whenever robots are expected to be of service to people \cite{bardaro2022robots}, as well as in the case of mixed teams where robots and humans co-operate towards shared goals} \cite{tiddi2023knowledge}. 

In this context, the model building problem can be seen as one of devising a unified representation of the environment that is both machine-readable and human-understandable. Thus, the concept of model building is highly interlinked with the long-standing issue of grounding observations and raw sensory information to symbols \cite{coradeschi2003introduction, bonarini2001anchoring}. In Robotics, efforts towards tackling this issue have consolidated the proposal of semantic maps, world models that contain ``in addition to spatial information about the environment, assignments of mapped features to entities of known classes'' \cite{nuchter2008towards}. As such, semantic maps provide an intermediate representation between sub-symbolic features learned directly from the environment and semantic symbols that can be mapped to natural language. Hence, as argued in \cite{chiatti2023visual}, semantic maps are a prime example of Neuro-symbolic system, which synthesises the inputs of different data-driven and knowledge-driven components. Ideally, semantic maps should not only be spatially contextualised but also reflect the evolution of the target environment, providing a spatio-temporal record of the robot's observations and activities (e.g., the set of waypoints and objects encountered along the robot's navigation route).  

In the remainder of this section, we explore a model building method based on semantic maps and instantiate this general method in assistive robotics scenarios to discuss the potential of this solution from the standpoint of mutual understanding.  

\subsection{Problem Scenario}
Brian is the primary caregiver for John, who is starting to show mild dementia symptoms. John often misplaces his copy of the house keys, and forgets where he has put his belongings, like his walking stick, or glasses. Brian decides to join a pilot study where commercial humanoid robots are deployed in selected households to help assisting elderly patients at home while their caregivers are away. Thus, this scenario concerns the interaction between human and robotic agents.   

On the day the robot is delivered to the house, the team of researchers running the study sets up the platform to map the entire layout of Brian's house. This 2D map will provide a basis for the robot to localise itself during future monitoring rounds. 

\begin{figure*}[t]
\includegraphics[width=\textwidth,clip]{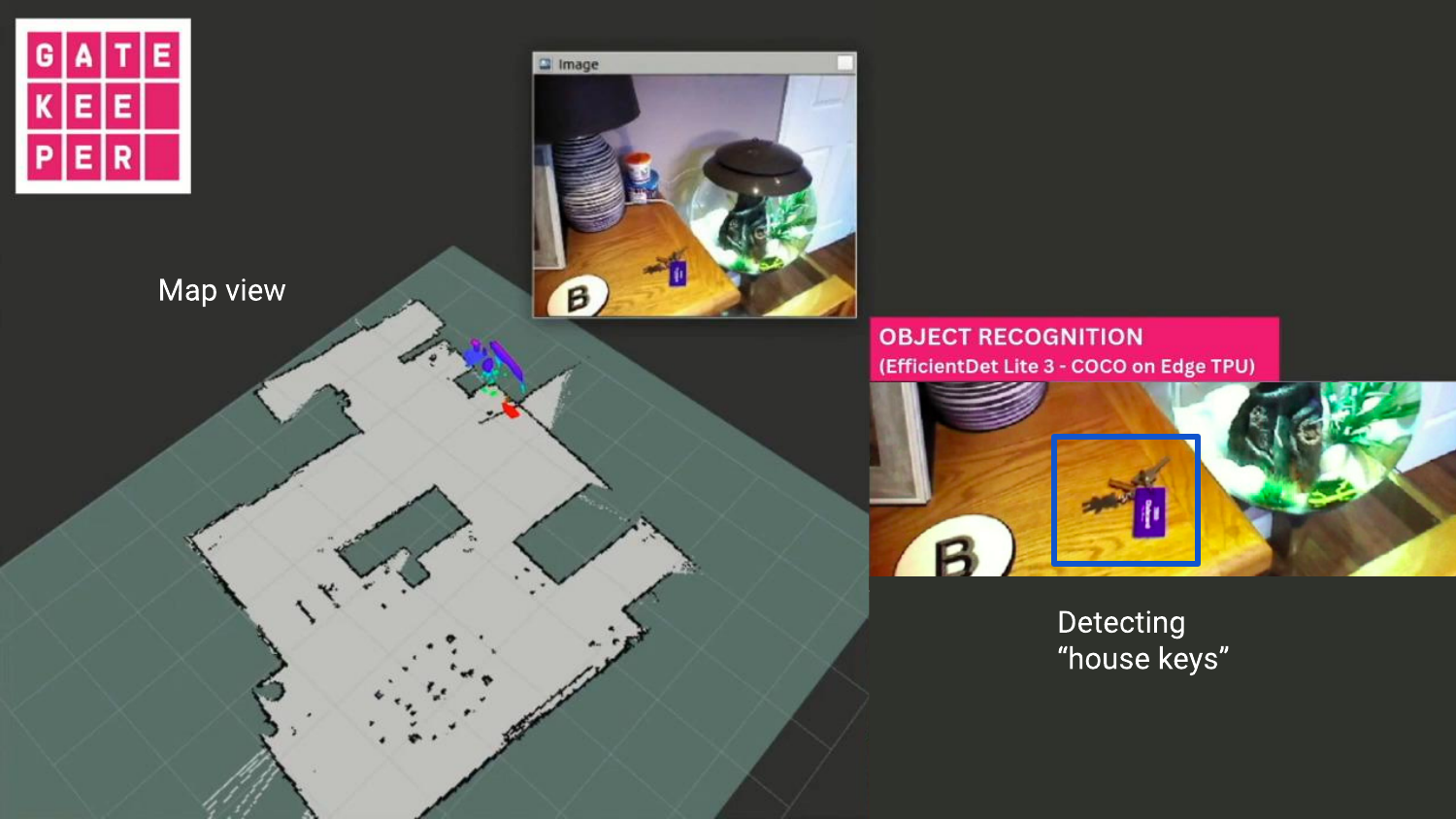}
  \caption{Visualization of the object retrieval routine. Picture courtesy of the Gatekeeper EU project consortium. On the left: the depth measurements collected through the robot camera (multi-color) are projected on the 2D house floorplan. On the right: the RGB image stream is fed to the object recognition module to detect the presence of personal items.}
  \label{fig:robot_at_home} 
\end{figure*}

The researchers would like to test how often the robot can accurately retrieve John's personal items when instructed to. During an initial setup phase, the user selects a few objects that the robot must learn to recognize. While patrolling, the robot identifies and records the locations of these items. Subsequently, the user can request the robot's assistance to retrieve a specific object. Since the tested solution is still under development, the robot simply locates the items and notifies the users without actually grasping any of the objects, to avoid damaging equipment and causing incidents. A sample visualization of the object retrieval routine is provided in Figure \ref{fig:robot_at_home}.   

\subsection{Technical Solution} 
To fit the requirements of a mobile robot that monitors the environment while navigating it, we developed a solution that comprises object recognition and spatial reasoning capabilities.
Importantly, the proposed framework is robust to variations in the robot's viewpoint and object orientation. Indeed, as the robot moves, its viewpoint changes over time. Moreover, the objects observed may follow different spatial orientations. Therefore, we introduce a \textit{contextualised frame of reference} \cite{chiatti2022spatial}, which allows to interpret the configuration of spatial objects with respect to both the robot's viewpoint and the location of nearby objects (e.g., ``car keys on the left of a potted plant"). In the proposed framework, geometric spatial operators are also mapped to common spatial predicates in natural language, to ensure that users' instructions (e.g., ``bring me the cup that is on the kitchen table") can be opportunely parsed and linked to the relevant geometric operations.  

At the core of this solution is the construction of a semantic map. That is, the 2D layout map acquired in the robot setup phase is enhanced with semantic information about the nature of the objects (e.g., a chair, a table), labels representing the different rooms (e.g., a kitchen, a bedroom), and the relationships between objects (e.g., a chair near the table, cable in the corridor). In practice, we represent object instances recognised through the robot's RGB-Depth (RGB-D) camera as anchors stored in a spatial database. Anchors serve as 3D abstractions of physical objects, formed by aggregating multiple measurements returned by the object recognition module. 

\begin{figure*}[t]
\includegraphics[width=\textwidth,trim=0 100 0 0, clip]{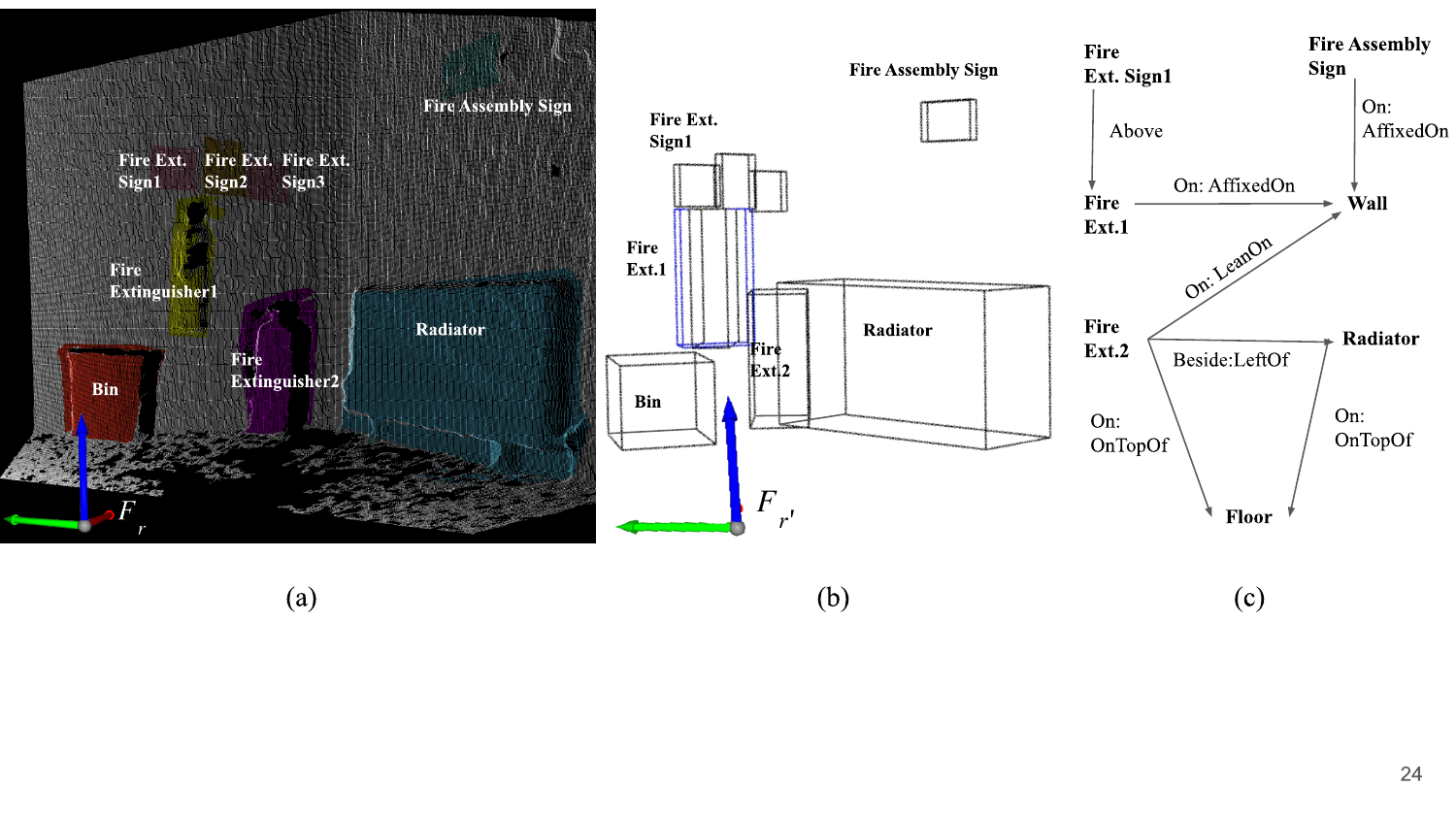}
  \caption{Example of operational workflow, taken from \cite{chiatti2022spatial}: (a) the PointCloud of the observed scene is segmented and annotated with object categories. Then, (b) the minimum oriented bounding boxes and contextualised bounding boxes (in blue) are constructed. Lastly, (c) a set of qualitative spatial relations is derived. }
  \label{fig:robot_mb_spatial} 
\end{figure*}

At each time frame, the distance between the robot's pose and the surfaces detected by the laser in the depth sensor is measured. These data, referred to as \textit{depth images}, can be converted into collections of 3D geometrical points within the given frame of reference, known as \textit{PointClouds}. Having access to Pointcloud measurements allows us to transform the 2D bounding boxes that represent object regions as minimum oriented 3D boxes, by applying the Convex Hull algorithm to the segmented PointCloud. We store the resulting spatial anchors in a PostgreSQL database, to capitalise on the availability of 3D spatial operators provided with the PostGIS engine and the SFCGAL backend. Thanks to these spatial operators, we can compute spatial relations between the detected objects.  A concrete example of the operational steps followed to extract spatial relations from raw RGB-D data is showcased in Figure \ref{fig:robot_mb_spatial}. Further details on the qualification process we followed to map PostGIS operators to spatial predicates expressed in natural language (e.g., \textit{onTopOf}, \textit{near}, \textit{affixedTo}, \textit{LeftOf}, etc.) can be found in \cite{chiatti2022spatial}. 

Anchors are intrinsically linked to the robot's perception. Thus, an anchor for a specific object does not exist unless the robot has detected that object. Similarly, an anchor must be removed from the semantic map if no new perceptions confirm its current location. The anchor life cycle is handled through automatic triggers in the PostgreSQL database. Namely, existing anchors are kept updated with the latest measurements that match the centroid location of the 3D bounding box. Whenever a new measurement is received that cannot be matched to any existing anchor, a new anchor is generated. 

For the object recognition module, we relied on Few-Shot Learning (FSL) methods that can learn from limited training examples, i.e., only the few pictures of the target personal items collected during the robot setup phase. Moreover, we prioritised lightweight models that could be fine-tuned on device via an Edge TPU USB-powered device, directly installed on the robot. 

The implemented solution provides a common world model that can be further extended even when the robot assistant is not the only device used for collecting environmental data. Smart Home sensors, for instance, can usefully complement the robot's observations \cite{bardaro2022robots}, e.g., enabling to infer the patient's location through a smart watch or fitness tracker. 
In this context, the semantic map provides a unified model to reconcile data collected from multiple sources and modalities. 

\subsection{Positioning w.r.t. Mutual Understanding Framework}
The model building problem is mainly aligned with the \textit{sharing knowledge} dimension of mutual understanding. Indeed, the semantic map introduced in this section acts as a common conceptual model that mediates between the robot's sensemaking, the data collection via external environmental sensors, and the users' requests. Through the semantic map, the robot and human operators can access the perceptual predictions produced by object detection models based on Deep Learning, as well as semantic relations between objects computed by the spatial engine.

Two specific interventions have been explored to achieve a common ground between the human and artificial actors involved in this scenario. First, we adopted a frame of reference that more clearly expresses the robot's viewpoint from which the relative positioning of objects is inferred (``keys on coffee table"). This configuration provides an alternative to directly relying on the coordinate system of the map (``keys North of main door"), which is not adequately intelligible to users. Second, we mapped the geometric spatial operators used to extract spatial relations from sensory data (e.g., \textit{3DIntersection}, \textit{Rotation}) to linguistic predicates commonly used in natural language to describe spatial relations between objects (e.g., \textit{near}, \textit{rightOf}).  

In sum, the proposed representational model provides a more transparent, easier-to-understand interface for: i) patients and caregivers interacting directly with the robot, ii) researchers assessing the robot's capability to comprehend the environment.  

\subsection{Open Challenges and Possible Extensions}
Accurate and reliable model building remains an open challenge and the aforementioned solution opens up many opportunities of extension. In the following, we focus primarily on those directions of improvement that directly impact the goal of achieving a mutual understanding between robots and users. 

The solution described in the previous section implies that whenever the user asks the robot to retrieve a specific object, only the last known location of the object is visited. If the object is not found, the full space is scouted to update the semantic map. This approach is sub-optimal from the standpoint of navigation, and it overlooks key information priors that can be derived from previous interactions with the user. For instance, users may repeatedly leave certain items at typical locations, which the robot could visit before scouting other areas of the house. This extension would help making more explicit the implicit user patterns and norms that contribute to mutual understanding. It would also offer an alternative direction of exploration to the increasing number of works focusing on goal-based navigation from a single image example \cite{li2022object}. These fully neural pipelines jointly learn map construction and symbol grounding via end-to-end optimisation. Hence, they limit feature reuse under changing environmental conditions \cite{hu2021architecture} and do not take advantage of previous knowledge of the monitored environment.      

Moreover, the proposed solution does not discriminate between object classes when handling anchor updates. However, objects like furniture and large appliances (sofa, fridge, TV stand) are less likely to change their location over time, while movable objects like personal items, chairs and utensils ought to be re-mapped more frequently. This is implicitly understood by us humans as we process the world visually. In fact, our assessment of previously-constructed mental models is only triggered in the presence of change, making our information processing most efficient \cite{snowden2012basic}. Another key extension contributing to building a common cognitive ground between robot assistants and humans may thus concern eliciting and integrating relevant commonsense knowledge representing the spatio-temporal trajectories of objects \cite{chiatti_towards_2020}. 

The extensions discussed above concern knowledge sharing in mutual understanding. From the standpoint of exchanging knowledge, the discussed solution provides a very limited coverage. In this configuration, the robot receives the user instructions and starts searching for a target object, without communicating its plans and activities back to the user. However, establishing a two-way interaction would significantly contribute to the mutual understanding of both involved actors, especially in cases where an adaptive switching between different tasks is required. The most recent advancement of solutions based on Large Language Models (LLMs) can significantly facilitate the integration of this missing link. Despite the limited ability of state-of-the-art LLMs to effectively manage robot planning tasks \cite{kambhampati2024can,kambhampati2024position}, they are undoubtedly powerful tools for translating and summarising programmed actions (i.e., code) to natural language. 

This improvement could be further extended by equipping assistive robots with the capability to infer patients' requests from implicit feedback. For example, in the motivating scenario of this section, the patient's expressed intention of ``going out for a walk" could trigger the robot's search for the walking stick. These implicit user preferences and patterns can be learned by directly interacting with the user and observing the environment, overcoming the burden of modelling explicitly user-specific constraints. Thus, this function is best supported by neural methods, such as Reinforcement Learning solutions~\cite{maroto2024personalizing}. Learning adaptive robot behaviors, in tandem with the extraction of typical object locations, directly contributes to consolidating the user's trust and acceptance of the system, and to the dimension of governing knowledge.

%% file: sezioni/hybrid-reasoning.tex
To operate reliably in real-world environments, autonomous robots ought to be equipped with advanced reasoning and decision-making capabilities. Having access to accurate world models, as discussed in the previous section, is only part of the equation: methods for advanced robot deliberation are also required \cite{ingrand2017deliberation}. Specifically, to ensure that a \textit{mutual understanding is reached between the autonomous agent and the users and personnel interacting with the robot, it is essential to ensure that these deliberation processes are handled in a controlled and transparent way}. 

This requirement is especially crucial in the case of safety-critical applications, such as in search and rescue operations, transportation, and medicine. In these challenging settings, the ability to learn and reason \textit{bottom-up} directly from the environment through information acquired at the sensory level needs to be opportunistically coupled with \textit{top-down} cognitive expectations about the environment \cite{cummings2021rethinking}. Namely, observations collected bottom-up via sub-symbolic learning ought to be combined with top-down symbolic knowledge modelling the environment, in line with the definition of Neuro-symbolic AI (NeSy) \cite{sarker2021neuro}. Indeed, as the complexity of the environment increases and in the presence of uncertainty, i.e., incomplete and conflicting information, lower-level pattern recognition and sensory-motor skills are insufficient for successful problem solving. Knowledge-based behaviors ought to emerge, where mental models built over time support the formulation and selection of specific plans \cite{cummings2021rethinking}. 

In the remainder of this section, we use the case of a robot assistant that monitors the Health and Safety of office environments as a running example to discuss the potential and open challenges of building Neuro-symbolic systems to support hybrid (top-down and bottom-up) reasoning. \\

\subsection{Problem Scenario}
The Health and Safety Robot Inspector (HanS) is expected to periodically patrol an office environment to spot any potential causes of harms to the Health and Safety of employees. Thus, similarly to the previous scenario, this use case involves the interaction between human and robotic agents to achieve shared goals. 

HanS ought to interpret the current state of the environment to identify any potential risks. First and foremost, this task entails robustly recognising objects. It also requires for HanS to be capable of considering the different properties that characterise the observed objects. Additionally, the robot should be aware of any constraints that govern the safety of the environment. 

Ultimately, given a set of observed objects with their related properties and a set of constraints, the robot must select the Health and Safety rules that are most relevant to the observed situation. If any violation to these rules is detected, HanS must promptly notify the designated fire wardens. 

\subsection{Technical Solution}
To address the problem of autonomously assessing the risk of an observed situation through a mobile robot, we developed the following modules. 

\paragraph{Semantic mapping} As in the previous use case, a semantic map is autonomously built representing the robot's operational environment. In this case, we extend the interface presented in the previous section to allow users to annotate Areas of Interest (AoI) on the map. For the HanS case study, we defined three types of AoI: (i) fire escape routes, (ii) waste collection areas, and (iii) areas representing the fire call points. Nonetheless, the same interface could be seamlessly applied to define other types of AoI, such as rooms, product-specific stocking spaces, or other landmarks relevant to the end application. The map (and associated spatial database) is populated with anchors generated via the following module.  

\begin{figure*}[t]
\includegraphics[width=\textwidth,trim=0 0 0 0, clip]{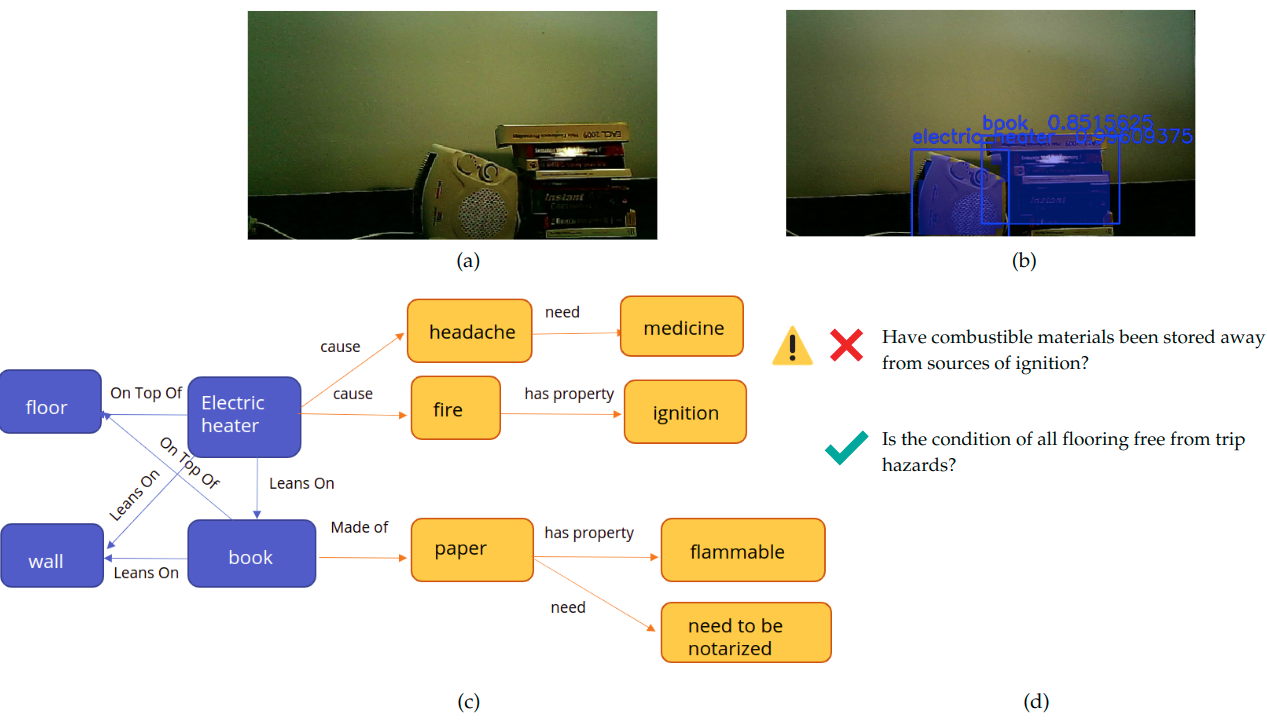}
  \caption{An electric heater is spotted near a pile of books. Because a flammable object is in contact with an ignition source, this scenario represents a potential fire hazard. }
  \label{fig:hans1} 
\end{figure*}

\paragraph{3D object anchoring.} Object predictions (i.e., their class and 2D bounding boxes) are predicted via Deep Learning from the RGB image stream collected through the robot camera (Figure\ref{fig:hans1}a). The generated bounding boxes are used to segment the raw PointClouds, yielding 3D object regions that can be stored in the spatial database, to compute the size and spatial relations of each object. Different observations are linked to the same object (i.e., anchor) if the distance between the centroids of their 3D boxes is lower than a pre-defined threshold, in terms of Euclidean distance.  

\paragraph{Size and spatial reasoning.} Based on the estimated size and spatial relations of each anchor, Deep Learning (DL) predictions selected as ``potentially incorrect" are autonomously corrected by relying on a supporting knowledge base. Specifically, we consider prior knowledge of the typical size and spatial relations of objects, repurposed from external knowledge bases (e.g., mugs are typically smaller than armchairs; fire extinguishers are usually affixed to the wall) to modify the ranking of predictions returned by the DL algorithm. Namely, we sourced typical sizes from ShapeNet \cite{savva2015semantically}, Amazon and manually-collected measurements, while typical spatial relations were extracted from Visual Genome \cite{krishna2017visual}. Further details on this phase can be found in \cite{chiatti2022robots}.   

\paragraph{Scene graph generation.} The consolidated object anchors and extracted spatial relations are used to construct a scene graph, i.e., a graph were nodes represent the objects in a scene and edges symbolise relationships between different objects. Relying on this representation allowed us to autonomously enrich the graph with additional symbolic knowledge expressing object properties as triplets (subject, predicate, object: e.g., \textit{paper is flammable}, Figure \ref{fig:hans1}c). Specifically, we retrieved the object fabrication materials (e.g., \textit{book made of paper}) from ShapeNet and other causal and material properties relevant to HanS' use-case from Quasimodo \cite{romero2020inside}. Crucially, the scene graph format enables the application of rule-based inference techniques, to verify any violations to the Health and Safety rules, as exemplified in Figure \ref{fig:hans1}d. To this aim, we formally express rules from the fire wardens' checklists and internal documents as First Order Logic (FOL) axioms. 

\subsection{Positioning w.r.t. Mutual Understanding Framework}
Similarly to the previous section, this case is rooted in the robot's model building capability and is thus closely linked with the dimension of \textit{sharing knowledge}. In addition to the semantic map, which acts as an overarching world model, scene graphs are also introduced as another key representation to describe successive fram sequences. However, the HanS use-case also makes an initial contribution to the two remaining ingredients of mutual understanding. 

Regarding the \textit{exchanging knowledge} dimension, the proposed NeSy framework capitalises on collective knowledge previously curated as large-scale knowledge repositories to: i) autonomously verify and correct the robot's predictions, generated via Deep Learning, ii) enrich the produced scene graphs with relevant contextual knowledge of the object properties and fabrication materials.  

Ultimately, the robot's risk assessment is regulated by a set of reference rules. We elicited these rules from standard practices followed by security officers and expressed them in a format that enables direct testing against the scene graphs. This step contributes to the challenge of \textit{governing knowledge} in mutual understanding.  

Overall, the type of knowledge sources exchanged and accessed in this use case scenario range from latent representations used for classifying objects through Deep Learning methods to structured Knowledge Bases (like Quasimodo, Visual Genome, and ShapeNet) and formal logic constraints describing safety rules. 

\subsection{Open Challenges and Possible Extensions}
The recent rise of Vision Language Models (VLMs) pre-trained on large-scale collections of images and text has expedited the progress on scene understanding and scene graph generation tasks \cite{wang2023cogvlm,chen2024spatialvlm}. Therefore, the discussed solution can be significantly updated by querying a VLM, instead of devoting two separate modules to object recognition and scene graph generation. From the standpoint of mutual understanding, dialoguing with a VLM would increase the users' participation in the robot's activities. For instance, Computer Vision tasks would be expressed in a conversational format (e.g., ``which objects do you see in this scene and what is their location?"). However, in this context, scene graphs remain an important representational tool to complement textual prompts. First, scene graphs embedded in prompts have been found to augment the compositional reasoning capabilities of VLMs on visual tasks \cite{mitra2024compositional}. Moreover, scene graphs can provide a graphical aid for users interacting with the robot to comprehend and verify the model responses, making the robot's deliberation functions more transparent.  

The Neuro-symbolic reasoning pipeline presented in the previous section, albeit leveraging top-down and bottom-up reasoning to infer object classes from images, operates only in one direction: background knowledge modelling the robot's environment guides the correction of the sub-symbolic predictions. However, a valuable feedback loop could be established, where the robot's bottom-up observation of the environment informs the generation of new knowledge, yielding a fully bi-directional NeSy pipeline. This improvement can especially facilitate the extraction of tacit knowledge.  Indeed, tacit knowledge is often taken for granted as we reach mutual understanding in everyday interactions, because it is mainly learned through experience and inherently difficult to express and formalise \cite{davis_commonsense_2015}.

%% file: main.bbl
\begin{thebibliography}{10}

\bibitem{akbar2018effects}
Fatema Akbar, Ted Grover, Gloria Mark, and Michelle~X Zhou.
\newblock The effects of virtual agents' characteristics on user impressions and language use.
\newblock In {\em Proceedings of the 23rd International Conference on Intelligent User Interfaces Companion}, page~56. ACM, 2018.

\bibitem{mubashara2024}
Mubashara Akhtar, Omar Benjelloun, Costanza Conforti, Pieter Gijsbers, Joan Giner-Miguelez, Nitisha Jain, Michael Kuchnik, Quentin Lhoest, Pierre Marcenac, Manil Maskey, Peter Mattson, Luis Oala, Pierre Ruyssen, Rajat Shinde, Elena Simperl, Goeffry Thomas, Slava Tykhonov, Joaquin Vanschoren, Jos van~der Velde, Steffen Vogler, and Carole-Jean Wu.
\newblock Croissant: A metadata format for ml-ready datasets.
\newblock In {\em Proceedings of the Eighth Workshop on Data Management for End-to-End Machine Learning}, DEEM '24, page 1–6, New York, NY, USA, 2024. Association for Computing Machinery.

\bibitem{tra2024governance}
Antonia Azzini, Marco Comerio, Sabino Metta, and Mario Scrocca.
\newblock The {TANGENT} governance model for mobility data sharing.
\newblock In {\em {Transport Transitions: Advancing Sustainable and Inclusive Mobility - Proceedings of the 10th TRA Conference}}, {Lecture Notes in Mobility}. Springer, 2024.

\bibitem{bardaro2022robots}
Gianluca Bardaro, Alessio Antonini, and Enrico Motta.
\newblock Robots for elderly care in the home: A landscape analysis and co-design toolkit.
\newblock {\em International Journal of Social Robotics}, 14(3):657--681, 2022.

\bibitem{belhajjame2014research}
Khalid Belhajjame, Jun Zhao, Daniel Garijo, Matthew Gamble, Kristina Hettne, Raul Palma, Eleni Mina, Oscar Corcho, José~Manuel Gómez-Pérez, Sean Bechhofer, Graham Klyne, and Carole Goble.
\newblock Using a suite of ontologies for preserving workflow-centric research objects.
\newblock {\em Journal of Web Semantics}, 32:16--42, 2015.

\bibitem{bellan2022extracting}
Patrizio Bellan, Mauro Dragoni, and Chiara Ghidini.
\newblock Extracting business process entities and relations from text using pre-trained language models and in-context learning.
\newblock In {\em International Conference on Enterprise Design, Operations, and Computing}, pages 182--199. Springer, 2022.

\bibitem{bonarini2001anchoring}
Andrea Bonarini, Matteo Matteucci, and Marcello Restelli.
\newblock Anchoring: do we need new solutions to an old problem or do we have old solutions for a new problem.
\newblock In {\em Proceedings of the AAAI Symposium on Anchoring Symbols to Sensor Data in Single and Multiple Robot Systems}, pages 79--86, 2001.

\bibitem{booch2021thinking}
Grady Booch, Francesco Fabiano, Lior Horesh, Kiran Kate, Jonathan Lenchner, Nick Linck, Andreas Loreggia, Keerthiram Murgesan, Nicholas Mattei, Francesca Rossi, et~al.
\newblock Thinking fast and slow in ai.
\newblock In {\em Proceedings of the AAAI Conference on Artificial Intelligence}, volume~35, pages 15042--15046, 2021.

\bibitem{breit_combining_2023}
Anna Breit, Laura Waltersdorfer, Fajar~J. Ekaputra, Sotirios Karampatakis, Tomasz Miksa, and Gregor Käfer.
\newblock Combining {Semantic} {Web} and {Machine} {Learning} for {Auditable} {Legal} {Key} {Element} {Extraction}.
\newblock In Catia Pesquita, Ernesto Jimenez-Ruiz, Jamie McCusker, Daniel Faria, Mauro Dragoni, Anastasia Dimou, Raphael Troncy, and Sven Hertling, editors, {\em The {Semantic} {Web}}, pages 609--624, Cham, 2023. Springer Nature Switzerland.

\bibitem{breit2023combining}
Anna Breit, Laura Waltersdorfer, Fajar~J Ekaputra, Marta Sabou, Andreas Ekelhart, Andreea Iana, Heiko Paulheim, Jan Portisch, Artem Revenko, Annette~ten Teije, et~al.
\newblock Combining machine learning and semantic web: A systematic mapping study.
\newblock {\em ACM Computing Surveys}, 55(14s):1--41, 2023.

\bibitem{cantone_ontological_2021}
Domenico Cantone, Carmelo~Fabio Longo, Marianna Nicolosi~Asmundo, Daniele~Francesco Santamaria, and Corrado Santoro.
\newblock Ontological {Smart} {Contracts} in {OASIS}: {Ontology} for {Agents}, {Systems}, and {Integration} of {Services}.
\newblock In {\em Intelligent {Distributed} {Computing} {XIV}}, Studies in {Computational} {Intelligence}, pages 237--247, Cham, 2022. Springer International Publishing.

\bibitem{carenini_enabling_2021}
Alessio Carenini, Andrea Fiano, Mario Scrocca, Marco Comerio, and Irene Celino.
\newblock Enabling {Cross}-{Border} {Travel} {Offers} {Through} {National} {Access} {Point} {Federation} via {Metadata} {Harmonisation}.
\newblock In David Chaves-Fraga, Pieter Colpaert, Mersedeh Sadeghi, Mario Scrocca, and Marco Comerio, editors, {\em Proceedings of the 3rd {International} {Workshop} {Semantics} {And} {The} {Web} {For} {Transport}}, volume 2939 of {\em {CEUR} {Workshop} {Proceedings}}, Online, September, September 2021. CEUR.
\newblock ISSN: 1613-0073.

\bibitem{carriero2024human}
Valentina~Anita Carriero, Antonia Azzini, Ilaria Baroni, Mario Scrocca, and Irene Celino.
\newblock Human evaluation of procedural knowledge graph extraction from text with large language models.
\newblock In Mehwish Alam, Marco Rospocher, Marieke van Erp, Laura Hollink, and Genet~Asefa Gesese, editors, {\em Proceedings of the 24th International Conference on Knowledge Engineering and Knowledge Management (EKAW 2024)}, pages 434--452, Cham, 2024. Springer Nature Switzerland.

\bibitem{celino2020coney}
Irene Celino and Gloria Re~Calegari.
\newblock Submitting surveys via a conversational interface: An evaluation of user acceptance and approach effectiveness.
\newblock {\em International Journal of Human-Computer Studies}, 139:102410, 2020.

\bibitem{celino2020refining}
Irene Celino, Gloria Re~Calegari, and Andrea Fiano.
\newblock Refining linked data with games with a purpose.
\newblock {\em Data Intelligence}, 2(3):417--442, 2020.

\bibitem{celino2021participant}
Irene Celino, Gloria Re~Calegari, Mario Scrocca, Jaime Zamorano, and Esteban Gonzalez~Guardia.
\newblock Participant motivation to engage in a citizen science campaign: the case of the tess network.
\newblock {\em Journal of Science Communication}, 20(6):A03, 2021.

\bibitem{chaves2024dagstuhlkg}
David Chaves-Fraga, Oscar Corcho, Anastasia Dimou, Maria-Esther Vidal, Ana Iglesias-Molina, and Dylan Van~Assche.
\newblock {Are Knowledge Graphs Ready for the Real World? Challenges and Perspective (Dagstuhl Seminar 24061)}.
\newblock {\em Dagstuhl Reports}, 14(2):1--70, 2024.

\bibitem{chen2024spatialvlm}
Boyuan Chen, Zhuo Xu, Sean Kirmani, Brain Ichter, Dorsa Sadigh, Leonidas Guibas, and Fei Xia.
\newblock Spatialvlm: Endowing vision-language models with spatial reasoning capabilities.
\newblock In {\em Proceedings of the IEEE/CVF Conference on Computer Vision and Pattern Recognition}, pages 14455--14465, 2024.

\bibitem{chiatti2023visual}
Agnese Chiatti, Gianluca Bardaro, Matteo Matteucci, and Enrico Motta.
\newblock Visual model building for robot sensemaking: Perspectives, challenges, and opportunities.
\newblock In {\em Proceedings of the Bridge Session on AI and Robotics of the thirty-seventh AAAI conference on Artificial Intelligence (AAAI-23), Washington, WA, USA}, pages 7--14, 2023.

\bibitem{chiatti2022spatial}
Agnese Chiatti, Gianluca Bardaro, Enrico Motta, and Enrico Daga.
\newblock A spatial reasoning framework for commonsense reasoning in visually intelligent agents.
\newblock In {\em Proceedings of the 8th International workshop on Artificial Intelligence and Cognition (AIC 2022)}. CEUR, 2022.

\bibitem{chiatti_towards_2020}
Agnese Chiatti, Enrico Motta, and Enrico Daga.
\newblock Towards a {framework} for {Visual} {Intelligence} in {Service} {Robotics}: {epistemic} {requirements} and {gap} {analysis}.
\newblock In {\em Proceedings of the International Conference on Principles of Knowldge Representation and Reasoning (KR) -{Special} session on {KR} \& {Robotics}}, pages 905--916. IJCAI, 2020.

\bibitem{chiatti2022robots}
Agnese Chiatti, Enrico Motta, and Enrico Daga.
\newblock Robots with commonsense: Improving object recognition through size and spatial awareness.
\newblock In {\em Proceedings of the AAAI Symposium on Machine Learning and Knowledge Engineering for Hybrid Intelligence (AAAI-MAKE)}. CEUR, 2022.

\bibitem{tra2024harmonisation}
Marco Comerio, Andrea Fiano, Marco Grassi, and Mario Scrocca.
\newblock Mobility data harmonisation: the {TANGENT} solution.
\newblock In {\em {Transport Transitions: Advancing Sustainable and Inclusive Mobility - Proceedings of the 10th TRA Conference}}, {Lecture Notes in Mobility}. Springer, 2024.

\bibitem{coradeschi2003introduction}
Silvia Coradeschi and Alessandro Saffiotti.
\newblock An introduction to the anchoring problem.
\newblock {\em Robotics and autonomous systems}, 43(2-3):85--96, 2003.

\bibitem{cummings2021rethinking}
Mary Cummings.
\newblock Rethinking the maturity of artificial intelligence in safety-critical settings.
\newblock {\em AI Magazine}, 42(1):6--15, 2021.

\bibitem{davis_commonsense_2015}
Ernest Davis and Gary Marcus.
\newblock Commonsense reasoning and commonsense knowledge in artificial intelligence.
\newblock {\em Communications of the ACM}, 58(9):92--103, 2015.

\bibitem{ding2021few}
Ning Ding, Guangwei Xu, Yulin Chen, Xiaobin Wang, Xu~Han, Pengjun Xie, Hai-Tao Zheng, and Zhiyuan Liu.
\newblock Few-nerd: A few-shot named entity recognition dataset.
\newblock {\em arXiv preprint arXiv:2105.07464}, 2021.

\bibitem{aiact}
{European Parliament} and {Council of the European Union}.
\newblock Regulation (eu) 2024/1689 of the european parliament and of the council of 13 june 2024 laying down harmonised rules on artificial intelligence, 2024.

\bibitem{Grassi2023Composable}
Marco Grassi, Mario Scrocca, Alessio Carenini, Marco Comerio, and Irene Celino.
\newblock Composable semantic data transformation pipelines with {Chimera}.
\newblock In {\em Proceedings of the 4th International Workshop on Knowledge Graph Construction co-located with 20th Extended Semantic Web Conference}, volume 3471 of {\em CEUR Workshop Proceedings}, Hersonissos, Greece, 5 2023. CEUR.
\newblock ISSN: 1613-0073.

\bibitem{guinan1986development}
Patricia Guinan and Robert~P Bostrom.
\newblock Development of computer-based information systems: A communication framework.
\newblock {\em ACM SIGMIS Database: the DATABASE for Advances in Information Systems}, 17(3):3--16, 1986.

\bibitem{hofer_towards_2024}
Marvin Hofer, Johannes Frey, and Erhard Rahm.
\newblock Towards self-configuring {Knowledge} {Graph} {Construction} {Pipelines} using {LLMs} - {A} {Case} {Study} with {RML}.
\newblock In David Chaves-Fraga, Anastasia Dimou, Ana Iglesias-Molina, Umutcan Serles, and Dylan~Van Assche, editors, {\em Proceedings of the 5th {International} {Workshop} on {Knowledge} {Graph} {Construction}}, volume 3718 of {\em {CEUR} {Workshop} {Proceedings}}, Hersonissos, Greece, May 2024. CEUR.
\newblock ISSN: 1613-0073.

\bibitem{hosseini_automated_2019}
Marjan Hosseini, Safia Kalwar, Matteo Rossi, and Mersedeh Sadeghi.
\newblock Automated {Mapping} for {Semantic}-based {Conversion} of {Transportation} {Data} {Formats}.
\newblock In Lucie-Aimee Kaffee, Kemele~M. Endris, Maria-Esther Vidal, Marco Comerio, Mersedeh Sadeghi, David Chaves-Fraga, and Pieter Colpaert, editors, {\em Joint {Proceedings} of the 1st {International} {Workshop} {On} {Semantics} {For} {Transport} and the 1st {International} {Workshop} on {Approaches} for {Making} {Data} {Interoperable}}, volume 2447 of {\em {CEUR} {Workshop} {Proceedings}}, Karlsruhe, Germany, September 2019. CEUR.
\newblock ISSN: 1613-0073.

\bibitem{hu2021architecture}
Jie Hu, Liujuan Cao, Tong Tong, Qixiang Ye, Shengchuan Zhang, Ke~Li, Feiyue Huang, Ling Shao, and Rongrong Ji.
\newblock Architecture disentanglement for deep neural networks.
\newblock In {\em Proceedings of the IEEE/CVF international conference on computer vision}, pages 672--681, 2021.

\bibitem{ingrand2017deliberation}
F{\'e}lix Ingrand and Malik Ghallab.
\newblock Deliberation for autonomous robots: A survey.
\newblock {\em Artificial Intelligence}, 247:10--44, 2017.

\bibitem{kahneman2011thinking}
Daniel Kahneman.
\newblock {\em Thinking, fast and slow}.
\newblock Farrar, Straus and Giroux, 2011.

\bibitem{kambhampati2024can}
Subbarao Kambhampati.
\newblock Can large language models reason and plan?
\newblock {\em Annals of the New York Academy of Sciences}, 1534(1):15--18, 2024.

\bibitem{kambhampati2024position}
Subbarao Kambhampati, Karthik Valmeekam, Lin Guan, Mudit Verma, Kaya Stechly, Siddhant Bhambri, Lucas~Paul Saldyt, and Anil~B Murthy.
\newblock Position: Llms can’t plan, but can help planning in llm-modulo frameworks.
\newblock In {\em Forty-first International Conference on Machine Learning}, 2024.

\bibitem{krishna2017visual}
Ranjay Krishna, Yuke Zhu, Oliver Groth, Justin Johnson, and et~al.
\newblock Visual genome: Connecting language and vision using crowdsourced dense image annotations.
\newblock {\em International Journal of Computer Vision}, 123(1):32--73, 2017.

\bibitem{kumar2020}
Abhijeet Kumar, Abhishek Pandey, Rohit Gadia, and Mridul Mishra.
\newblock Building {Knowledge} {Graph} using {Pre}-trained {Language} {Model} for {Learning} {Entity}-aware {Relationships}.
\newblock In {\em 2020 {IEEE} {International} {Conference} on {Computing}, {Power} and {Communication} {Technologies} ({GUCON})}, pages 310--315, 2020.

\bibitem{lake_building_2017}
Brenden~M. Lake, Tomer~D. Ullman, Joshua~B. Tenenbaum, and Samuel~J. Gershman.
\newblock Building machines that learn and think like people.
\newblock {\em Behavioral and Brain Sciences}, 40, 2017.

\bibitem{law2011human}
Edith Law and Luis Von~Ahn.
\newblock {\em Human computation}.
\newblock Morgan \& Claypool Publishers, 2011.

\bibitem{lewis2020retrieval}
Patrick Lewis, Ethan Perez, Aleksandra Piktus, Fabio Petroni, Vladimir Karpukhin, Naman Goyal, Heinrich K{\"u}ttler, Mike Lewis, Wen-tau Yih, Tim Rockt{\"a}schel, et~al.
\newblock Retrieval-augmented generation for knowledge-intensive nlp tasks.
\newblock {\em Advances in Neural Information Processing Systems}, 33:9459--9474, 2020.

\bibitem{li2022object}
Baosheng Li, Jishui Han, Yuan Cheng, Chong Tan, Peng Qi, Jianping Zhang, and Xiaolei Li.
\newblock Object goal navigation in eobodied ai: A survey.
\newblock In {\em Proceedings of the 2022 4th International Conference on Video, Signal and Image Processing}, pages 87--92, 2022.

\bibitem{maroto2024personalizing}
Marcos Maroto-G{\'o}mez, Mar{\'\i}a Malfaz, Jos{\'e}~Carlos Castillo, {\'A}lvaro Castro-Gonz{\'a}lez, and Miguel~{\'A}ngel Salichs.
\newblock Personalizing activity selection in assistive social robots from explicit and implicit user feedback.
\newblock {\em International Journal of Social Robotics}, pages 1--19, 2024.

\bibitem{mitra2024compositional}
Chancharik Mitra, Brandon Huang, Trevor Darrell, and Roei Herzig.
\newblock Compositional chain-of-thought prompting for large multimodal models.
\newblock In {\em Proceedings of the IEEE/CVF Conference on Computer Vision and Pattern Recognition}, pages 14420--14431, 2024.

\bibitem{montgomery1981form}
Barbara~M Montgomery.
\newblock The form and function of quality communication in marriage.
\newblock {\em Family Relations}, pages 21--30, 1981.

\bibitem{nuchter2008towards}
Andreas N{\"u}chter and Joachim Hertzberg.
\newblock Towards semantic maps for mobile robots.
\newblock {\em Robotics and Autonomous Systems}, 56(11):915--926, 2008.

\bibitem{randles_r2rml-chatgpt_2024}
Alex Randles and Declan O’Sullivan.
\newblock R2[{RML}]-{ChatGPT} {Framework}.
\newblock In David Chaves-Fraga, Anastasia Dimou, Ana Iglesias-Molina, Umutcan Serles, and Dylan~Van Assche, editors, {\em Proceedings of the 5th {International} {Workshop} on {Knowledge} {Graph} {Construction}}, volume 3718 of {\em {CEUR} {Workshop} {Proceedings}}, Hersonissos, Greece, May 2024. CEUR.
\newblock ISSN: 1613-0073.

\bibitem{nikitha2024}
Nikitha Rao, Jason Tsay, Kiran Kate, Vincent Hellendoorn, and Martin Hirzel.
\newblock Ai for low-code for ai.
\newblock In {\em Proceedings of the 29th International Conference on Intelligent User Interfaces}, IUI '24, page 837–852, New York, NY, USA, 2024. Association for Computing Machinery.

\bibitem{re2018interplay}
Gloria Re~Calegari and Irene Celino.
\newblock Interplay of game incentives, player profiles and task difficulty in games with a purpose.
\newblock In {\em Knowledge Engineering and Knowledge Management: 21st International Conference, EKAW 2018, Nancy, France, November 12-16, 2018, Proceedings 21}, pages 306--321. Springer, 2018.

\bibitem{re2018framework}
Gloria Re~Calegari, Andrea Fiano, and Irene Celino.
\newblock A framework to build games with a purpose for linked data refinement.
\newblock In {\em The Semantic Web--ISWC 2018: 17th International Semantic Web Conference, Monterey, CA, USA, October 8--12, 2018, Proceedings, Part II 17}, pages 154--169. Springer, 2018.

\bibitem{re2018human}
Gloria Re~Calegari, Gioele Nasi, and Irene Celino.
\newblock Human computation vs. machine learning: an experimental comparison for image classification.
\newblock {\em Human Computation}, 5:13--30, 2018.

\bibitem{romero2020inside}
Julien Romero and Simon Razniewski.
\newblock Inside quasimodo: Exploring construction and usage of commonsense knowledge.
\newblock In {\em Proceedings of the 29th ACM International Conference on Information \& Knowledge Management}, pages 3445--3448, 2020.

\bibitem{rula2023procedural}
Anisa Rula and Jennifer D'Souza.
\newblock Procedural text mining with large language models.
\newblock In {\em Proceedings of the 12th Knowledge Capture Conference 2023}, pages 9--16, 2023.

\bibitem{rula2023annotation}
Anisa Rula, Gloria Re~Calegari, Antonia Azzini, Ilaria Baroni, and Irene Celino.
\newblock Annotation and extraction of industrial procedural knowledge from textual documents.
\newblock In {\em Proceedings of the 12th Knowledge Capture Conference 2023}, pages 1--8, 2023.

\bibitem{rula2022eliciting}
Anisa Rula, Gloria Re~Calegari, Antonia Azzini, Davide Bucci, Ilaria Baroni, and Irene Celino.
\newblock Eliciting and curating procedural knowledge in industry: Challenges and opportunities.
\newblock {\em Qurator}, 2022.

\bibitem{rula2023khubonto}
Anisa Rula, Gloria Re~Calegari, Antonia Azzini, Davide Bucci, Alessio Carenini, Ilaria Baroni, and Irene Celino.
\newblock K-hub: a modular ontology to support document retrieval and knowledge extraction in industry 5.0.
\newblock In {\em European Semantic Web Conference}, pages 454--470. Springer, 2023.

\bibitem{sadeghi_interoperability_2024}
Mersedeh Sadeghi, Alessio Carenini, Oscar Corcho, Matteo Rossi, Riccardo Santoro, and Andreas Vogelsang.
\newblock Interoperability of heterogeneous {Systems} of {Systems}: from requirements to a reference architecture.
\newblock {\em The Journal of Supercomputing}, 80(7):8954--8987, May 2024.

\bibitem{sarker2021neuro}
Md~Kamruzzaman Sarker, Lu~Zhou, Aaron Eberhart, and Pascal Hitzler.
\newblock Neuro-symbolic artificial intelligence.
\newblock {\em AI Communications}, 34(3):197--209, 2021.

\bibitem{savva2015semantically}
Manolis Savva, Angel~X Chang, and Pat Hanrahan.
\newblock Semantically enriched 3d models for common-sense knowledge.
\newblock In {\em Proceedings of the IEEE Conference on Computer Vision and Pattern Recognition Workshops (CVPRW)}, pages 24--31. IEEE, 6 2015.

\bibitem{scrocca2024integration}
Mario Scrocca, Ilaria Baroni, Alessio Carenini, Marco Comerio, and Irene Celino.
\newblock {An Integration Framework to Support Data Sharing and Process Tracking in Intelligent Asset Management Systems}.
\newblock In {\em Proceedings of the 10th {Transport} {Research} {Arena}}, {Lecture Notes in Mobility}. Springer, 2024.

\bibitem{scrocca2024not}
Mario Scrocca, Alessio Carenini, Marco Grassi, Marco Comerio, and Irene Celino.
\newblock Not everybody speaks rdf: Knowledge conversion between different data representations.
\newblock In {\em Fifth International Workshop on Knowledge Graph Construction@ ESWC2024}, 2024.

\bibitem{Scrocca2020Turning}
Mario Scrocca, Marco Comerio, Alessio Carenini, and Irene Celino.
\newblock Turning transport data to comply with {EU} standards while enabling a multimodal transport knowledge graph.
\newblock In {\em Proceedings of the 19th International Semantic Web Conference}, volume 12507, page 411–429. Springer, 2020.

\bibitem{ride2rail}
Mario Scrocca, Marco Comerio, Alessio Carenini, and Irene Celino.
\newblock Modelling {Business} {Agreements} in the {Multimodal} {Transportation} {Domain} {Through} {Ontological} {Smart} {Contracts}.
\newblock In {\em Towards a {Knowledge}-{Aware} {AI}}, pages 137--151. IOS Press, 2022.

\bibitem{scrocca2024intelligent}
Mario Scrocca, Marco Grassi, Marco Comerio, Valentina~Anita Carriero, Tiago {Delgado Dias}, Ana {Vieira Da Silva}, and Irene Celino.
\newblock Intelligent urban traffic management via semantic interoperability.
\newblock In Gianluca Demartini, Katja Hose, Maribel Acosta, Matteo Palmonari, Gong Cheng, Hala Skaf-Molli, Nicolas Ferranti, Daniel Hern{\'a}ndez, and Aidan Hogan, editors, {\em Proceedings of the 23rd International Semantic Web Conference (ISWC2024)}, pages 218--235, Cham, 2024. Springer Nature Switzerland.

\bibitem{scrocca2021survey}
Mario Scrocca, Damiano Scandolari, Gloria {Re Calegari}, Ilaria Baroni, and Irene Celino.
\newblock {The Survey Ontology: Packaging Survey Research as Research Objects}.
\newblock In {\em Proceedings of the 2nd Workshop on Data and Research Objects Management for Linked Open Science (DaMaLOS), co-located with ISWC 2021}, 2021.

\bibitem{shi2024generative}
Senbao Shi, Zhenran Xu, Baotian Hu, and Min Zhang.
\newblock Generative multimodal entity linking.
\newblock In {\em Proceedings of the 2024 Joint International Conference on Computational Linguistics, Language Resources and Evaluation (LREC-COLING 2024)}, pages 7654--7665, 2024.

\bibitem{snowden2012basic}
Robert Snowden, Robert~J Snowden, Peter Thompson, and Tom Troscianko.
\newblock {\em Basic vision: an introduction to visual perception}.
\newblock Oxford University Press, 2012.

\bibitem{rocrate}
Stian Soiland-Reyes et~al.
\newblock Packaging research artefacts with {RO}-{Crate}.
\newblock {\em Data Science}, 5(2):97--138, January 2022.
\newblock Publisher: IOS Press.

\bibitem{stolk2016conceptual}
Arjen Stolk, Lennart Verhagen, and Ivan Toni.
\newblock Conceptual alignment: How brains achieve mutual understanding.
\newblock {\em Trends in cognitive sciences}, 20(3):180--191, 2016.

\bibitem{tan1994establishing}
Margaret Tan.
\newblock Establishing mutual understanding in systems design: An empirical study.
\newblock {\em Journal of Management Information Systems}, 10(4):159--182, 1994.

\bibitem{tan1989investigation}
Margaret~JY Tan.
\newblock An investigation of the communication behaviors of systems analysts.
\newblock {\em School of Business, The University of Queensland}, 1989.

\bibitem{tiddi2023knowledge}
Ilaria Tiddi, Victor De~Boer, Stefan Schlobach, and Andr{\'e} Meyer-Vitali.
\newblock Knowledge engineering for hybrid intelligence.
\newblock In {\em Proceedings of the 12th Knowledge Capture Conference 2023}, pages 75--82, 2023.

\bibitem{tsaneva2024enhancing}
Stefani Tsaneva and Marta Sabou.
\newblock Enhancing human-in-the-loop ontology curation results through task design.
\newblock {\em ACM Journal of Data and Information Quality}, 16(1):1--25, 2024.

\bibitem{van2021modular}
Michael van Bekkum, Maaike de~Boer, Frank van Harmelen, Andr{\'e} Meyer-Vitali, and Annette~ten Teije.
\newblock Modular design patterns for hybrid learning and reasoning systems: a taxonomy, patterns and use cases.
\newblock {\em Applied Intelligence}, 51(9):6528--6546, 2021.

\bibitem{vanharmelen2019boxology}
Frank Van~Harmelen and Annette Ten~Teije.
\newblock A boxology of design patterns for hybrid learning and reasoning systems.
\newblock {\em Journal of Web Engineering}, 18(1-3):97--123, 2019.

\bibitem{DCAT-AP}
Bert Van~Nuffelen.
\newblock {DCAT Application profile for data portals in Europe (DCAT-AP)}.
\newblock [Online; accessed 2024-07-16].

\bibitem{Vetere2005Models}
G.~Vetere and M.~Lenzerini.
\newblock Models for semantic interoperability in service-oriented architectures.
\newblock {\em IBM Systems Journal}, 44(4):887--903, 2005.

\bibitem{vonahn2006games}
Luis Von~Ahn.
\newblock Games with a purpose.
\newblock {\em Computer}, 39(6):92--94, 2006.

\bibitem{wang2023cogvlm}
Weihan Wang, Qingsong Lv, Wenmeng Yu, Wenyi Hong, Ji~Qi, Yan Wang, Junhui Ji, Zhuoyi Yang, Lei Zhao, Xixuan Song, et~al.
\newblock Cogvlm: Visual expert for pretrained language models.
\newblock {\em arXiv preprint arXiv:2311.03079}, 2023.

\bibitem{wiemann1977explication}
John~M Wiemann.
\newblock Explication and test of a model of communicative competence.
\newblock {\em Human communication research}, 3(3):195--213, 1977.

\bibitem{wittgenstein2009philosophical}
Ludwig Wittgenstein.
\newblock {\em Philosophical investigations}.
\newblock John Wiley \& Sons, 2009.

\bibitem{process-run-crate}
{Workflow Run RO-Crate Working Group}.
\newblock {Workflow Run RO-Crate Profiles - Process Run Crate}.

\bibitem{xiao2019tell}
Ziang Xiao, Michelle~X Zhou, Q~Vera Liao, Gloria Mark, Changyan Chi, Wenxi Chen, and Huahai Yang.
\newblock Tell me about yourself: Using an ai-powered chatbot to conduct conversational surveys.
\newblock {\em arXiv preprint arXiv:1905.10700}, 2019.

\bibitem{zhou2019trusting}
Michelle~X Zhou, Gloria Mark, Jingyi Li, and Huahai Yang.
\newblock Trusting virtual agents: The effect of personality.
\newblock {\em ACM Transactions on Interactive Intelligent Systems (TiiS)}, 9(2-3):10, 2019.

\end{thebibliography}
